\documentclass{article}

\usepackage{arxiv}

\usepackage[utf8]{inputenc} 
\usepackage[T1]{fontenc}    
\usepackage{hyperref}       
\usepackage{url}            
\usepackage{booktabs}       
\usepackage{amsfonts}       
\usepackage{nicefrac}       
\usepackage{microtype}      
\usepackage{lipsum}		
\usepackage{graphicx}
\usepackage{natbib}
\usepackage{doi}

\usepackage{verbatim}
\usepackage{graphicx}
\usepackage{url} 
\usepackage{listings}

\usepackage{subfigure}
\usepackage{wrapfig}
\usepackage{float}

\usepackage{amsmath}
\usepackage{amssymb}

\usepackage{algorithm}
\usepackage{algpseudocode}

\usepackage{pgfgantt}

\title{An Event based Prediction Suffix Tree}

\author{ \href{https://orcid.org/0009-0001-1920-5925}{\includegraphics[scale=0.06]{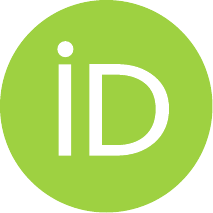}\hspace{1mm}Evie ~Andrew} \\
International Centre for Neuromorphic Systems (ICNS) \\
The MARCS Institute of Brain, Behaviour and Development \\
Western Sydney University \\
	\texttt{e.andrew@westernsydney.edu.au} \\
 	\And
 \href{https://orcid.org/0000-0002-9025-0198}{\includegraphics[scale=0.06]{orcid.pdf}\hspace{1mm}Travis ~Monk} \\
International Centre for Neuromorphic Systems (ICNS) \\
The MARCS Institute of Brain, Behaviour and Development \\
Western Sydney University \\
	\texttt{t.monk@westernsydney.edu.au} \\
	\And
 \href{https://orcid.org/0000-0001-6140-017X}{\includegraphics[scale=0.06]{orcid.pdf}\hspace{1mm}André van ~Schaik} \\
International Centre for Neuromorphic Systems (ICNS) \\
The MARCS Institute of Brain, Behaviour and Development \\
Western Sydney University \\
	\texttt{a.vanschaik@westernsydney.edu.au} \\
}

\renewcommand{\shorttitle}{An Event based Prediction Suffix Tree}

\hypersetup{
pdftitle={An Event based Prediction Suffix Tree},
pdfsubject={cs.NE},
pdfauthor={Evie ~Andrew, Travis ~Monk, André van ~Schaik},
pdfkeywords={Event Based, Sparse Coding, Neuromorphic Algorithm, Suffix Tree},
}

\begin{document}
\maketitle

\begin{abstract}
	This article introduces the Event based Prediction Suffix Tree (EPST), a biologically inspired, event-based prediction algorithm. The EPST learns a model online based on the statistics of an event based input and can make predictions over multiple overlapping patterns. The EPST uses a representation specific to event based data, defined as a portion of the power set of event subsequences within a short context window. It is explainable, and possesses many promising properties such as fault tolerance, resistance to event noise, as well as the capability for one-shot learning. The computational features of the EPST are examined in a synthetic data prediction task with additive event noise, event jitter, and dropout. The resulting algorithm outputs predicted projections for the near term future of the signal, which may be applied to tasks such as event based anomaly detection or pattern recognition.
\end{abstract}


\section{Introduction}
Many prediction algorithms only accept sequential data as input. The Hidden Markov Model (HMM) tracking algorithms  \cite{bayes_filtering}\cite{variational_bayes}\cite{VFFriston}, are designed for application to trajectories in n-dimensional space. Meanwhile, Variable order Markov Model (VMM) algorithms \cite{vmmpst}\cite{moffat1990PPMC}, exclusively handle sequences defined on a finite discrete state space.  In a biological context, neurons communicate via discrete events known as spikes, which are neither trajectories in n-dimensional space nor categorical data. As a result, event based data requires additional processing to become compatible with these sequential data formats. 

Event based data (or spikes) consist of sequences of discrete events at specific times across multiple channels. The study of event based data is a relatively new field compared to sequential data, despite being present in multiple practical applications, such as software information logging, network packet capture data, and event based sensors. 

There are multiple ways to interpret raw event data or transform it such that it is suitable for input into algorithms not originally designed to handle such data directly. In the neuromorphic literature, this is often referred to as a `representation' of a spike train. A comprehensive review of spike based representations is given in \cite{codingreview}, and a more recent review focuses on the robustness of representations to event noise \cite{spikecodereview1}. Most representations abstract a time interval containing spikes into a single floating point value, or a vector of them. Examples include the firing rate, time to first spike \cite{codingreview}, or a time surface \cite{spikecodetimesurfacesindexsurfaces}.

The neuromorphic engineering tool Nengo, which is based on the Neural Engineering Framework \cite{nef}\cite{nefbook}, also operates on the idea that a population's spike train can be abstracted to a floating point vector or value via a rate code or a phase code. Den\`eve \cite{deneve1} \cite{deneve2} uses instantaneous rates with a low pass filter to quantify the relative belief between two hypotheses. As discussed in the review \cite{codingreview}, rate codes are thought now to be too slow and information inefficient to be the principal form of communication in the biological brain.

Another common abstraction of event based data is the conversion to symbolic sequence data, or in other words an assignment of a symbol from a finite discrete set. This can be achieved via a Winner Take All (WTA) for `one-hot' encoding, or k-Winners Take All (kWTA) for more dense codes. Time To First Spike (TTFS) is a form of a Winner Take All representation where the first neuron to spike within a given interval represents a stimulus. Time To First Spike requires some agreed upon start time such as a stimulus presentation, and an end time for a code. Also, many implementations enforce simplifications such as a single spike per channel per time window \cite{spikecodettfs}. Phase codes are neuromorphic in the sense that they rely on regular rhythms to define the start and the end of a code, and the code itself may be compared to a binary word from computer science. A binary word is a fixed length sequence of 1s and 0s, and the positioning of these symbols within a binary word is important to its meaning. Binary codes like this have been extensively studied for their capacity, speed of interpretation, and robustness \cite{spikecodephase}\cite{spikecodenofm}. Methods that enforce synchronisation to a global clock are less scalable than a fully asynchronous solution. Furthermore, complex global information sharing is not biologically plausible, and each computational unit should only communicate locally with other neurons via spikes.

Rank order codes are favoured in the review \cite{codingreview} due to their theoretical capacity, however, in practical or uncontrolled settings, codes can decay easily in the presence of event noise (See Section \ref{EPSTRandomInterferenceSection}). Time surfaces and index surfaces can represent order based codes and carry approximate timing information efficiently in practical settings \cite{spikecodetimesurfaces} \cite{spikecodetimesurfacesindexsurfaces}. However, as in other codes, often a single spike per channel per interval must be enforced due to the lack of memory in any given channel for more than a single spike. A desirable processing algorithm would allow more than a single spike per channel to carry information. It would support simultaneous spikes, and be robust to the interaction of complex data with noise. Scenarios such as untangling multiple overlapping codes or extracting signal from noise with missing information are common scenarios encountered in practice and must be addressed.

Kernel projection methods \cite{tempotron}\cite{spikecodeskim2013}\cite{spikecodeskan2015}\cite{sinkernelseliasmith} are an interesting class of spike interpretation methods where a set of kernels are applied to spatiotemporal events defined on a moving time window. These kernel methods can define any shape depending on the specificity of the task and are particularly useful for classification tasks similar to the SVM \cite{cervantesSVM2020}. Training arbitrary kernels online in a neuromorphic way may be impossible, however, specific cases have been solved for fast online training \cite{spikecodeopium2015}. In unsupervised feature extraction, the FEAST algorithm allows adaptation of a neuron population's receptive fields and efficiently distributes these receptive fields throughout an input space \cite{spikecodefeast2020}. More recent work provides convenient ways to learn complex kernels using deep learning architectures \cite{spatiotemporalrepdeeplearning}. Of particular relevance, SKIM \cite{spikecodeskim2013}, illustrates the detection of a specific pattern in noise, and SKAN \cite{spikecodeskan2015} will select and become sensitive to a single arbitrary pattern that appears most frequently in a dataset. Kernel projection methods with decision boundaries often struggle with adversarial examples \cite{adversarial}, and as they tend to require iterative refinement in training, struggle with data efficiency. Other iterative refinement procedures such as gradient descent also tend to have low data efficiency, therefore, developing methods with higher data efficiency would be preferable. For instance, demonstrating recall of spike patterns after a single presentation in specific parameterisations should be sufficient.

The novel contribution of this paper is the exploration of a new method which attempts to address the issues of asynchronous action, global data sharing, data efficiency, fault tolerance, and robustness to noise. Many of these desired properties have been reported in the neuromorphic computation literature for VMMs and sequential symbolic data \cite{JeffH}. The Prediction Suffix Tree (PST) is a VMM and has been identified as an opportunity for adaptation to an event based version of the algorithm named the EPST. This process is outlined in Section \ref{PSTtoEPST}. The definition of the VMM prediction problem adapted to events as well as the implementation details are given in Sections \ref{MathsProblemDefinition} and \ref{implementationdetails}.
The resulting EPST algorithm defines a prediction model based directly on the timings and channels of events, allowing simultaneous events, and multiple spikes per channel per time window. This, and other interesting computational properties are explored in Section \ref{VMMComparisonChapter} using a synthetic dataset, with a comparison to VMMs that use a sequential interpretation of the data.

\section{Adapting the PST Algorithm for Event Based Data.}
\label{PSTtoEPST}

Here we adapt the PST algorithm to accept event based data as input. With some adjustments to the PST data structure, we can create a neuromorphic algorithm that processes spike events, as opposed to the symbolic strings handled natively by PSTs and Suffix Trees.

\begin{figure}
	\centering
	\includegraphics[width=0.9\textwidth]{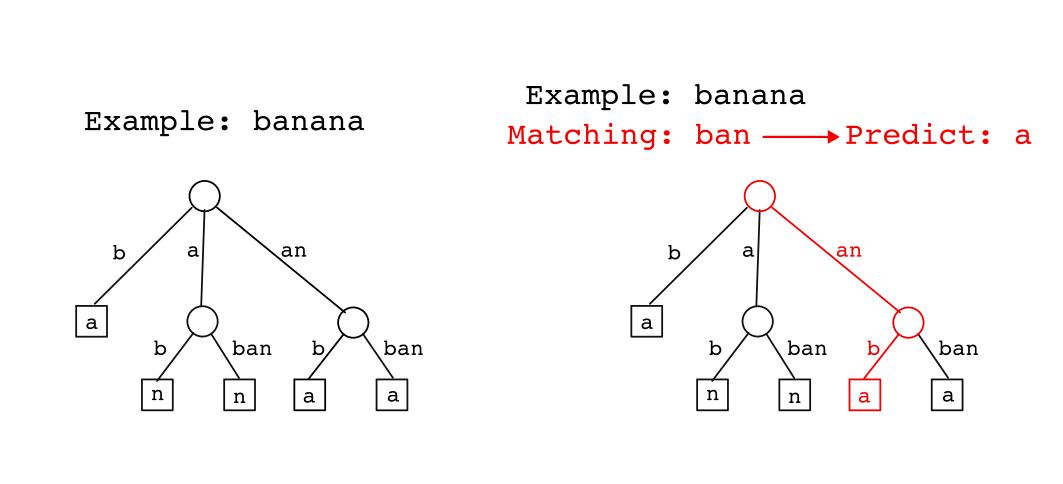}
	\caption{\label{EPSTtoPST1} \textbf{A diagram of a simplified suffix tree.} The left panel shows a simplified example of a sequential symbolic Suffix Tree, applied to the example sequence `banana'. The red line in the right panel of the figure illustrates how to traverse the tree given the suffix `ban' to predict the symbol `a'.}
\end{figure}

Figure \ref{EPSTtoPST1} shows an example of a sequential symbolic suffix tree, applied to the example sequence `banana'. It is built via the suffix tree algorithm \cite{vmmpst}, but for simplicity, we have excluded the calculation of probabilities. In a suffix tree, the leaf nodes contain the symbol predicted by the sequence stored along the branch. For example; the beginning symbol `b' predicts `a', `ba' predicts `n' and so does `bana', and so on. In red, Figure \ref{EPSTtoPST1} illustrates how to traverse the tree given the suffix `ban' to predict the symbol `a'.

\begin{figure}
	\centering
	\includegraphics[width=0.9\textwidth]{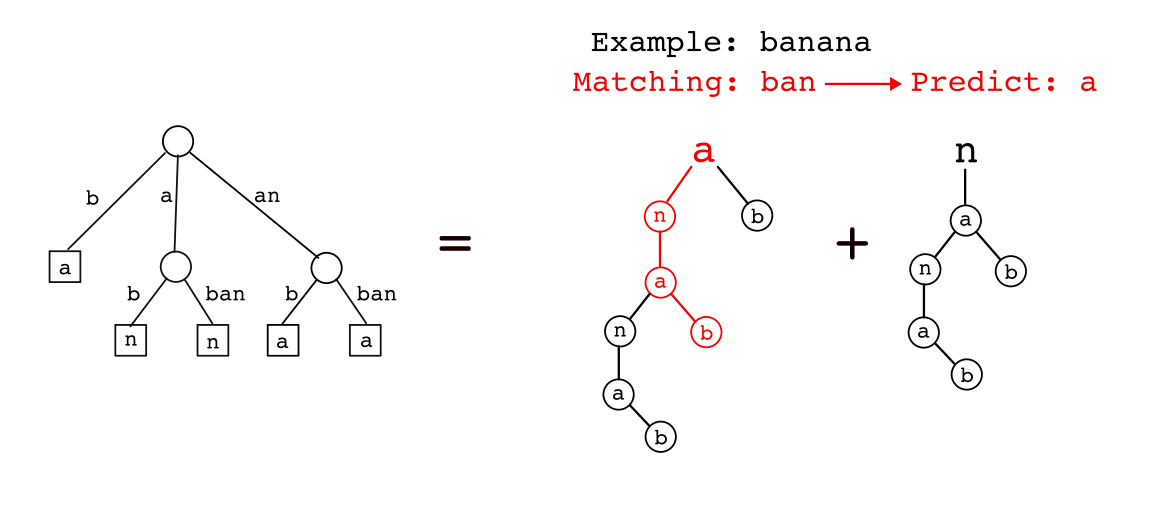}
	\caption{\label{EPSTtoPST2}  \textbf{A diagram of the decomposition of a suffix tree data structure to specialised trees.} Shows the first step to transforming the Suffix tree into the novel EPST, which is first to split the tree into smaller trees, each handling its own predicted symbol. The suffix tree specialised to predict `b' was omitted as it contains no prefixes. Since all branches of each new fragmented tree share a predicted symbol, we place this symbol at the root of the tree for notational simplicity. The red line again illustrates how to traverse the tree given the suffix `ban' to predict the symbol `a'. Note that none of the subsequences contained within the branches of the specialised `n' suffix tree match the context.}
\end{figure}

Figure \ref{EPSTtoPST2} shows the first step to transforming the suffix tree into the EPST. We split the tree into smaller trees, each handling its own predicted symbol. In this example the symbols `n' and `a' are present in many of the leaf nodes, so we can split them into their specialised suffix tree. All smaller trees collectively function identically to the original Suffix Tree. Figure \ref{EPSTtoPST2} once again in red illustrates how to traverse the tree given the suffix `ban' to predict the symbol `a'. This creates some abstract neuron like units, which in computational terms may be searched independently in parallel and even stored in different memory locations.

\begin{figure}
	\centering
	\includegraphics[width=0.9\textwidth]{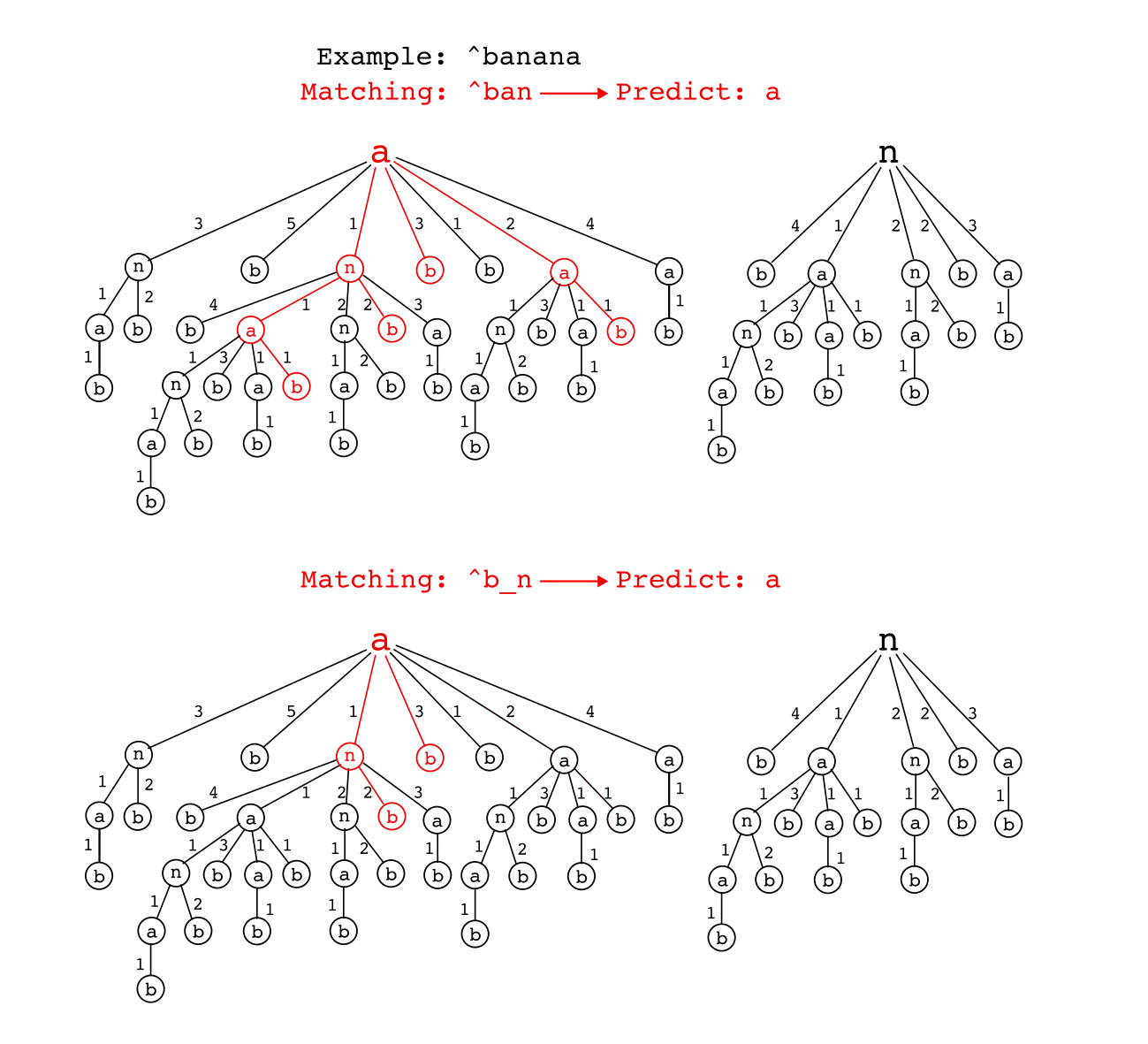}
	\caption{\label{EPSTtoPST3} \textbf{Adding delays to tree edges introduces redundancy and fault tolerance to a suffix tree.} Shows the final transformation required to create the EPST. `Time delay' values are represented on the edges of the tree, and are derived from the number of positions difference between elements in the example sequence `banana'. The resulting trees have also been adjusted to contain all combinations of subsequences from the suffixes in the example sequence. This is illustrated by an example in the top panel, where from the suffix `ban' to predict the symbol `a' we have `$b \stackrel{1}{\to} a \stackrel{1}{\to} n \stackrel{1}{\to} a$', `$b \stackrel{2}{\to} n \stackrel{1}{\to} a$', `$b \stackrel{1}{\to} a \stackrel{2}{\to} a$', `$a \stackrel{1}{\to} n \stackrel{1}{\to} a$', `$b \stackrel{3}{\to} a$', `$a \stackrel{2}{\to} a$', and `$n \stackrel{1}{\to} a$' with appropriate specific delays. The lower panel illustrates the EPSTs robustness to missing data. When presented with the broken subsequence `b\_n', the EPST can provide the correct predicted symbol `a', as highlighted in red in the diagram.}
\end{figure}

Figure \ref{EPSTtoPST3} shows the final transformation required to create the EPST. Since the EPST is event based, symbols from multiple channels may arrive simultaneously, and time steps may pass without an event being received, both scenarios not encountered in sequential data. As a result, the trees' edges must now contain delay information between events. Now that the suffix tree accepts event based data, we must address instances where the sequence contains noise events or other sequences. For example, the algorithm must recognise a sequence `$b \stackrel{2}{\to} a \stackrel{3}{\to} a$' with delays 2, 3 respectively as containing the sequence `$b \stackrel{5}{\to} a$' with delay 5 as a subset. While this is not an issue at all for simple sequence based data, it turns out to be a non-trivial problem for event based data. Of the many workarounds we could have chosen for the EPST, storing the power set of all subsequences and delays was deemed the simplest to study, leading to the large branching structure pictured in Figure \ref{EPSTtoPST3}. Finally in red, the top of Figure \ref{EPSTtoPST3} shows again how to traverse the tree given the suffix `ban' to predict the symbol `a', this differs from Figures \ref{EPSTtoPST1} and \ref{EPSTtoPST2} as now there are four redundant routes to the predicted symbol `a'. This redundancy provides robustness to additive noise and missing symbols, for example, the suffix `b\_n' missing a symbol `a' will also generate a correct prediction, as shown in the bottom of Figure \ref{EPSTtoPST3}.

\begin{figure}
	\centering
	\includegraphics[width=0.9\textwidth]{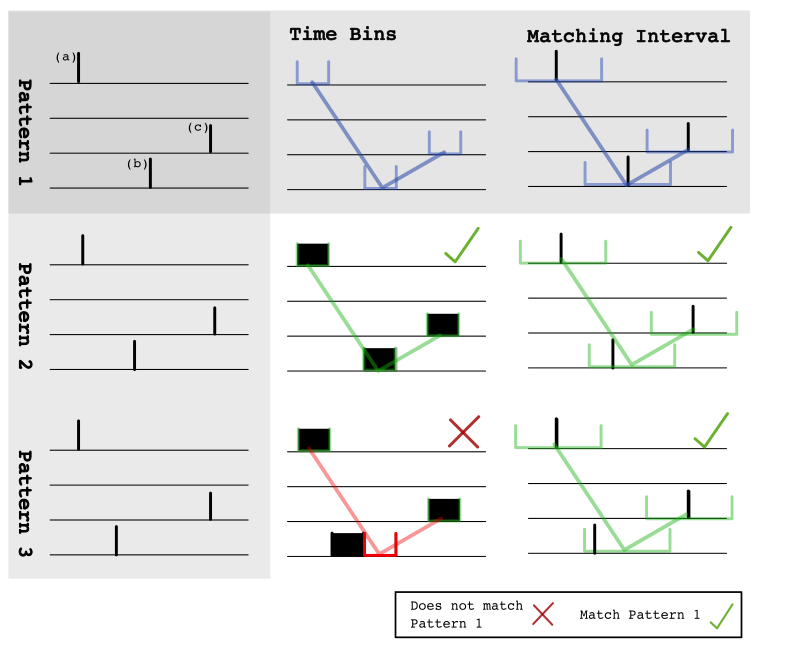}
	\caption{\label{time_matching_methods} \textbf{A schematic of methods for matching stored sequences with incoming spike patterns.}The left column shows 3 repetitions of pattern 1 with slight variations to event (b). The middle column shows the time bin method for matching subsequences. The top cell of this column shows the stored representation of pattern 1, note that each interval is grid aligned and the original exact timings of the events are not stored. For the time bin representation, pattern 3 is sufficiently different to not match the stored representation. The right column shows the matching interval method for matching subsequences. The matching interval method does store the exact timings of the events of pattern 1 and uses these timings to match subsequent patterns. An additional advantage of the matching interval method is a configurable sensitivity to variations in the timings of events, this is illustrated as both patterns 2 and 3 match the increased interval size.}
\end{figure}

Since an event theoretically has no duration, a matching rule must be devised to allow stored patterns to be matched with similar input patterns. The EPST may quantise time into small time bins, however, this does not function well if event times of equivalent patterns have random variations. Another more flexible solution is to define a region of time around each stored spike for which an incoming spike is considered equal. The distinction between matching methods is explored in Figure \ref{time_matching_methods}.

\section{Problem Definition}
\label{MathsProblemDefinition}

\begin{figure}
	\centering
	\includegraphics[width=0.8\textwidth]{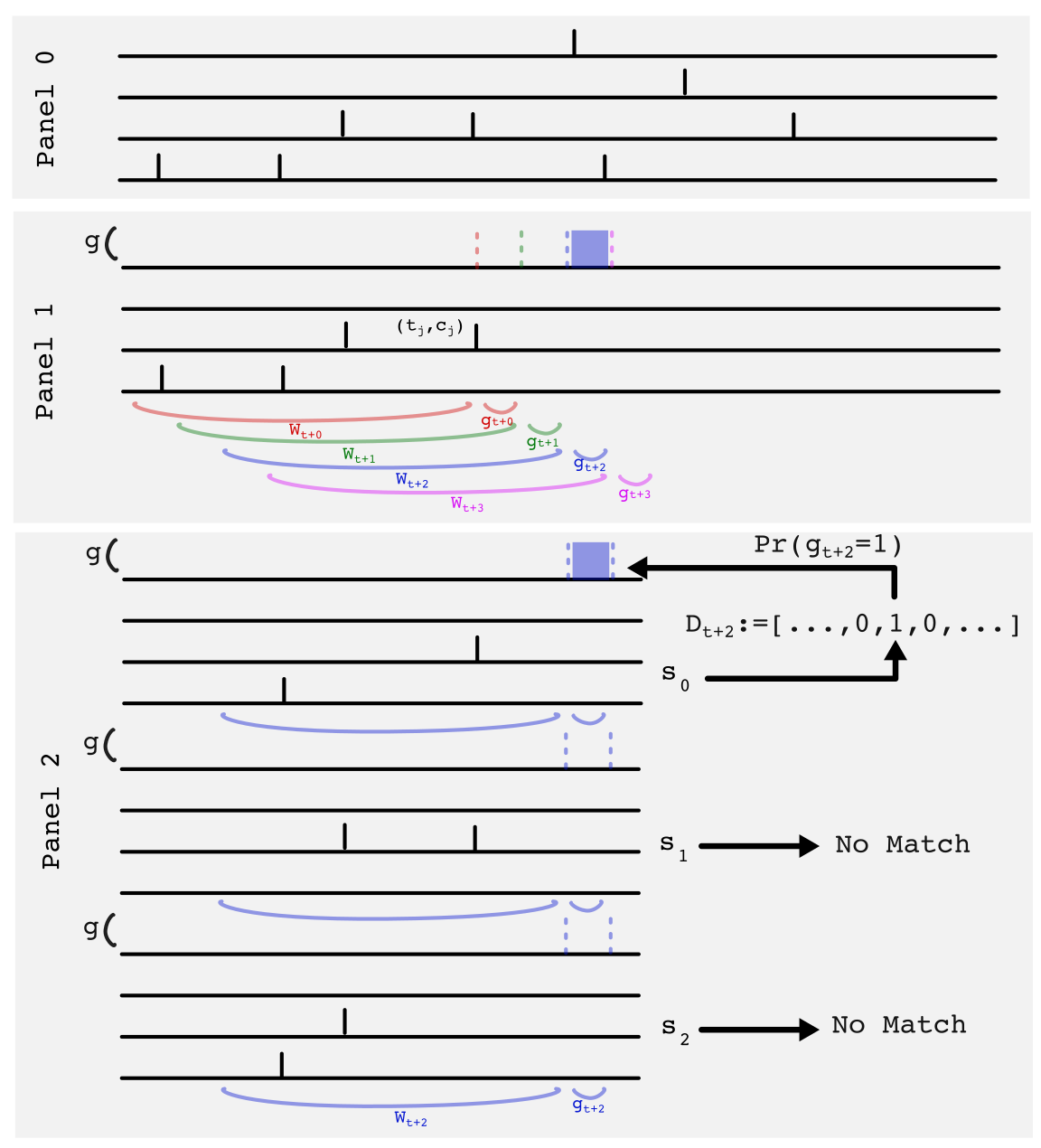}
	\caption{\label{EPSTDefinitionDiagram} \textbf{The EPST predicts the presence of spikes in a channel at a time, given the spatiotemporal patterns of past spikes.} Panel 0 shows an example spatiotemporal event sequence. Panel 1 shows a spike driven prediction triggered by the event $(t_{j},c_{j})$, at simulation time $t := t_{j}$. Each colour represents a step of the simulation for which the EPST predicts $g_{t+n}$ using features derived from the time window $W_{t+n}$. Panel 2 illustrates the feature extraction process from the time window $W_{t+2}$ to the binary feature vector $D_{t+2}$ and the resulting prediction of $g_{t+2}$. $s_{0}$, $s_{1}$, and $s_{2}$ are subsequences, or a subset of elements from the power set decomposition of the events in $W_{t+2}$. Subsequences that do not match are discarded (As an example, $s_{1}$ and $s_{2}$ do not have a corresponding entry in the feature vector $D$.), and matching subsequences ($s_{0}$) act as a key into a specific position within the $D_{t+2}$ feature vector, flipping the boolean to 1. The completed $D_{t+2}$ allows the calculation of the estimated probability for $g_{t+2}=1$. Details of probability estimates and building the vector $D$ are given in Section \ref{implementationdetails}.}
\end{figure}

The EPST problem is similar to the sequence prediction algorithm solved by VMM. However, since the EPST is event based this problem must be redefined, the steps detailed in this section are illustrated in Figure \ref{EPSTDefinitionDiagram}. 

Let $f$ be some spatiotemporal event sequence representing an outside stimulus (Figure \ref{EPSTDefinitionDiagram}, panel 0):

\[
f := \left[(t_{0}, c_{0}), (t_{1}, c_{1}), (t_{2}, c_{2}), ... , (t_{j}, c_{j})\right],  t_{j} \in \mathbb{R}, c_{j} \in \{0, ... , C\},
\]

where a spike event is a tuple $(t_{i}, c_{i})$ containing the time and channel of the $i^{th}$ spike, and $C$ is the number of channels to categorise the spike events. We assume that $t := t_{j}$ is the current simulation time.

The EPST is spike triggered in the sense that prediction (and therefore computation) is only triggered on receipt of a spike (Figure \ref{EPSTDefinitionDiagram}, panel 1). Spike triggered computation is intended to pave the way for more efficient computing solutions, as minimal computation is wasted in the space between spikes. Let $n$ index time steps following receipt of a spike at time $t$, or time bins $(t + n\delta t,t + (n + 1)\delta t], n \in \{0, ..., M'\}$. Here $M'$ is defined as the size of the prediction time window. 

The EPST calculates features from a history window $W_t$. Let $W_t$ be defined as the history of the signal preceding $t$:
\[
W_{t} := \{(t - t_{k}, c_{k})\}, \forall k,\text{ where }t_{k} \in \left[t-M,t\right),
\]

and $M$ is defined as the size of the history window. Note that the history window $W_{t}$ only contains the precise timing of spikes relative to the simulation time $t$, and contains no absolute timing information. Let $F$ be a lossy mapping of $W_{t+n}$ onto a binary feature vector of length $N_{D}$:

\[
F: W_{t} \to D_{t},\text{ where }D_{t} := \{0, 1\}_{N_{D} \times 1}.
\]

This mapping will represent the dictionary lookup table of subsequences in the EPST implementation.

Finally, let $g \in \{0, ... , C\}$ be an event channel `preferred' by an EPST unit. This is the channel that the EPST unit will learn to predict. Let $g_{t+n}:=1$ when an event exists within the time period $(t + n\delta t,t + n \delta t + \delta t]$ in the channel $g$, otherwise let $g_{t+n}:=0$. Note that $g$ may be any single channel event stream chosen by the EPST unit, and not only from the input channels $\{0, ... , C\}$, for example, when including a teaching signal.

Now, the goal of a single EPST unit is to find an estimated probability $\hat{Pr}(g_{t+n}=1)$, for each future timestep $n \in \{0, ..., M'\}$ based on the feature vector $D_{t+n}$. This process is repeated whenever a new input spike arrives.

The feature function $F$ is a power set decomposition of the events in $W_{t}$, illustrated in Figure \ref{EPSTDefinitionDiagram}, panel 2. All subsequences $s_{i} \in PowerSet(W_{t})$ are used as keys into the feature vector $D_{t}$, and we define $D_{t}(s_{i}) := 1$ if there is a match, and $D_{t}(s_{i}) := 0$ otherwise. $F$ has the Markov property in the same way as the VMM as it is only dependent on the contents of the relative history window $W_{t}$, and not on the absolute time $t$. 

\begin{figure}
	\centering
	\includegraphics[width=1.0\textwidth]{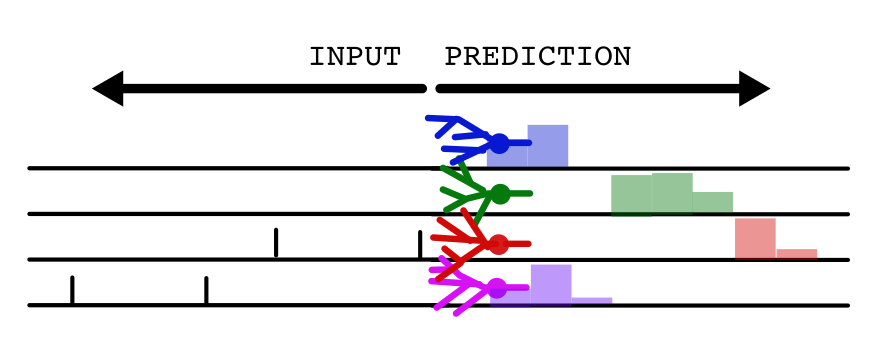}
	\caption{\label{EPSTMultiple} \textbf{A high level schematic diagram of multiple EPST units working in parallel} Each individual EPST functions as described in Figure \ref{EPSTDefinitionDiagram}, however, each predicts spikes in its assigned channel $g$. Each EPST unit receives spikes from all channels, but calculates predictions only for its assigned $g$. The left half of the figure shows the relative event history providing context for predictions. The right half of the figure shows bar graphs representing probability estimates for $\hat{Pr}(g_{t+n}=1)$ for each $g$, and each future timestep in the simulation window $n \in \{0, ..., M'\}$.}
\end{figure}

Figure \ref{EPSTMultiple} illustrates how multiple EPST units may be set up to predict many channels, each with a different channel assigned as their preference $g$. 

\section{Implementation Details}
\label{implementationdetails}

In Section \ref{MathsProblemDefinition} we introduced the EPST problem, which is a variation on the VMM prediction problem adapted to receive and predict event based data. Figure \ref{EPSTDefinitionDiagram} and \ref{EPSTMultiple} showed an overview of the prediction step of the EPST algorithm, defining the input from the relative history window $W_{t}$, the intermediate feature vector $D_{t}$, and the resulting predictions for each $g$. 

To complete the specification of the EPST algorithm, we must provide details of how the EPST learns the feature vector $D$ and calculates the estimates for $\hat{Pr}(g=1)$. This learning step is both spike triggered, and can be implemented entirely online using only information local to each EPST unit. In the following Section \ref{EPSTLearning} we will first describe the implementation of $D$ as an efficient tree data structure which facilitates the efficient collection and lookup of subsequences. Second, in Section \ref{EPSTProbabilityCalculation} we will show how the EPST collects statistics describing the relation between stored subsequences and spikes in $g$, and derive a formula to estimate probabilities for $g$ using these statistics. Finally, in Section \ref{EPSTRepresentativePattern} we describe a protocol for choosing one subsequence when many candidates are present in a particular relative history.

\subsection{Training the EPST by counting occurrences and growing the tree}
\label{EPSTLearning}

\begin{figure}
	\centering
	\includegraphics[width=0.9\textwidth]{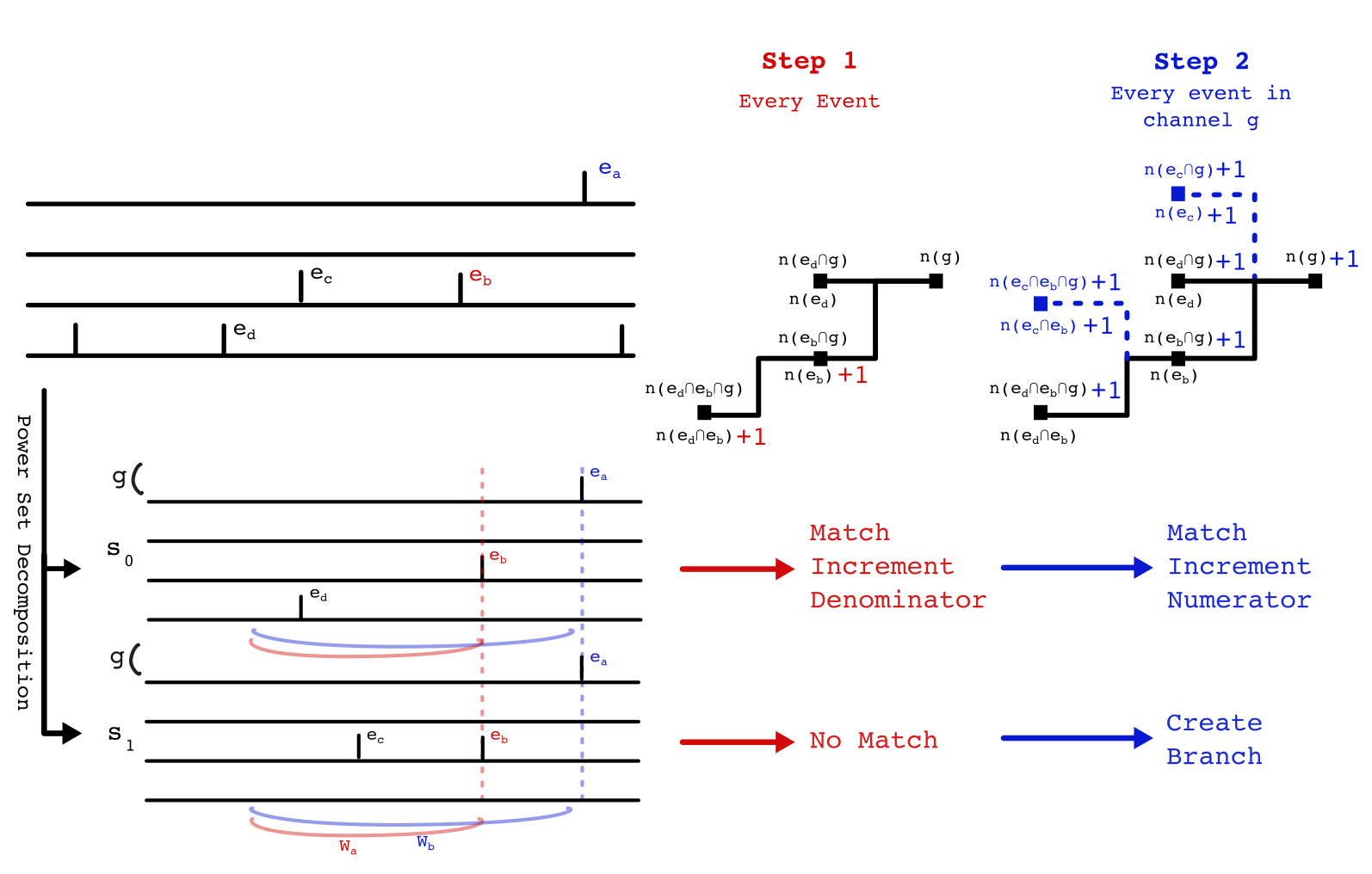}
	\caption{\label{EPSTLearningDiagram} \textbf{A detailed diagram illustrating the EPST learning steps.} The neuron in this diagram is assigned $g := c_{a}$ as its preferred channel, which is contained within the event tuple with the same name ($e_{a} := (t_{a}, c_{a})$). All associations made are relative, so the neuron is not aware of absolute time. Since the full pattern of spikes within the history of the spike in channel 0 may be rare (3 spikes pictured within the learning window $W_{t}, t:=t_{a}$), we decompose this full set into subsequences pictured ($s_{0}$ and $s_{1}$) using a power set operation. The power set produces many more than just these subsequences, however, they were omitted from the diagram for simplicity. These subsequences are expanded and used for each of the pictured learning steps 1 and 2. The data structure for the EPST is a tree, with counts stored at each node. Each node apart from the root holds two counts, in the diagram the top count is the numerator, and the bottom count is the denominator. Step 1 is triggered on receipt of any spike and increments the denominator of matching nodes, this example shows only step 1 triggered with respect to event $e_{b}$. Step 2 is triggered only upon receipt of the preferred event $g = c_{a}$. It increments the numerator of matching nodes, and also extends the tree to include more nodes under certain conditions.}
\end{figure}

As described in Section \ref{MathsProblemDefinition}, the feature map $F$ and feature vector $D_{t}$ function together as a look-up table of key value pairs, the key being a subsequence feature $s_{i}$, and the value being a binary digit indicating the presence of the feature. The feature look up table can be implemented efficiently as a tree data structure shown in Figure \ref{EPSTLearningDiagram}. This structure is memory efficient as subsequences can share suffixes with other similar subsequences, and subsequence matching can be implemented faster with a tree search.

The feature tree grows during training with the addition of new nodes, as shown in step 2 of Figure \ref{EPSTLearningDiagram}. The feature tree may also shrink through pruning, which is discussed in Section \ref{EPSTTreeSizeProblem}. Using the feature tree to create probability estimates requires counts to be collected about each subsequence and their relation to spikes in the EPSTs preferred channel $g$. Figure \ref{EPSTLearningDiagram} also demonstrates how statistics, stored as counts, are distributed among the nodes of the tree data structure.

Figure \ref{EPSTLearningDiagram} provides a simple example showing how counts in the tree are incremented, and in which conditions the increment is triggered. The notation $n(condition)$ will be used going forward to represent such counts. Since for all prediction times $t \in \{t_{T} + 0,...,t_{T} + M'\}$ any subsequence $s_{i}$ in the feature tree can either be present ($s_{i}=1$) or absent ($s_{i}=0$), and a spike in the preferred channel may either be present ($g_{t} = 1$) or absent ($g_{t} = 0$). The quantities sufficient to calculate estimates of probabilities are $n(s_{i} = 1), \forall i$, and $n(s_{i}=1\cap g_{t}=1), \forall i$. Below, the quantity $n(s_{i} = 1)$ is accumulated via step 1, and $n(s_{i}=1\cap g_{t}=1)$ is accumulated via step 2.

\textbf{Step 1} (Figure \ref{EPSTLearningDiagram}) is triggered on receipt of any spike in any channel. Explaining step 1 requires the definition of an `active top level subtree'. A `top level subtree' is a subtree with a root node that is a first level branch. What makes this subtree `active' is that the root event of the tree shares a channel with the received event. Now we can describe the conditions under which the denominator count of a node is incremented:
\begin{enumerate}
    \item {The node is a member of an active top level subtree.}
    \item {The subsequence defined between the node and the root of the active top level subtree is a match to the window $W_{t}$.}
\end{enumerate}
For example, from Figure \ref{EPSTLearningDiagram}, if the event that triggered step 1 is $e_{b} := (t_{b},c_{b})$, then the top level subtree with root node $e_{b'} := (t_{b'},c_{b'})$ is considered active if $c_{b} = c_{b'}$. Now another node within this top level subtree is $e_{d'}$ such that $e_{b'} := (t_{d'},c_{d'}),(t_{b'},c_{b'})$. The denominator of this node is incremented if there exists within $W_{b}$ an event $(t_{d},c_{d})$ such that $t - t_{d} \approx t_{b'} - t_{d'}$ and $c_{d} = c_{d'}$.

\textbf{Step 2} (Figure \ref{EPSTLearningDiagram}) is triggered only on receipt of a spike in channel $g$. The root count $n(g)$ of the tree is incremented, as well as the numerator counts of any nodes that match subsequences defined along paths starting from the root of the tree. Branches are extended from a node during step 2 if the numerator assigned to the node is greater than a hyperparameter called the `extend threshold' (See Section \ref{EPSTTreeSizeProblem}). This hyperparameter is analogous to a learning rate, and the higher it is, the slower the tree will grow.

\subsection{Estimating a probability using the EPST tree structure}
\label{EPSTProbabilityCalculation}

Finally, counts can be used to estimate a conditional probability with the simple formula:

\begin{align}
    \hat{Pr}(g_{t+n}=1 \cap s_{i}=1) &= \frac{n(g_{t+n}=1 \cap s_{i}=1)}{t} \label{numerator}\\
    \hat{Pr}(s_{i}=1) &= \frac{n(s_{i}=1)}{t} \label{denominator}\\
    \hat{Pr}(g_{t+n}=1|s_{i}=1) &= \frac{\hat{Pr}(g_{t+n}=1 \cap s_{i}=1)}{\hat{Pr}(s_{i}=1)} = \frac{n(g_{t+n}=1 \cap s_{i}=1)}{n(s_{i}=1)} \label{SimpleConditional}
\end{align}

Calculating a probability using the new tree data structure now only involves information stored locally in a particular node of the tree corresponding to the pattern match. 

In the following, the history sequence consists of the spikes $s = [(d_0, c_0), (d_1, c_1), ... (d_r, c_r)]$ where $r$ is the $r^{th}$ spike in increasing delay order measured from the present time $t$, with channel $c_r$ and delay $d_r$. Spike delays are strictly $d_0 \leq d_1 \leq ... d_r$ and so absolute spike times are $t - d_0 \geq t - d_1 \geq ... \geq t - d_r$. The probability of a later spike $e_{t, d_{r}, c_{r}}$, given an earlier spike $e_{t, d_{r-1}, c_{r-1}}$ is:

\begin{align*}
	\hat{Pr}(e_{t, d_{r}, c_{r}} = 1|e_{t, d_{r-1}, c_{r-1}} = 1) &= \frac{\hat{Pr}(e_{t, d_{r}, c_{r}} = 1 \cap e_{t, d_{r-1}, c_{r-1}} = 1)}{\hat{Pr}(e_{t, d_{r-1}, c_{r-1}} = 1)} \\
	\intertext{From empirical statistics we have:}
	\hat{Pr}(e_{t, d_{r}, c_{r}} = 1\cap e_{t, d_{r-1}, c_{r-1}} = 1) &= \frac{n(e_{t, d_{r}, c_{r}} = 1\cap e_{t, d_{r-1}, c_{r-1}} = 1)}{t}\\
        \hat{Pr}(e_{t, d_{r-1}, c_{r-1}} = 1) &= \frac{n(e_{t, d_{r-1}, c_{r-1}} = 1)}{t}\\
        \intertext{So we can estimate the conditional probability with counts:}
        \hat{Pr}(e_{t, d_{r}, c_{r}} = 1|e_{t, d_{r-1}, c_{r-1}} = 1) &= \frac{n(e_{t, d_{r}, c_{r}} = 1 \cap e_{t, d_{r-1}, c_{r-1}} = 1)}{n(e_{t, d_{r-1}, c_{r-1}} = 1)} \\
	\intertext{Similarly, we can represent arbitrary conditionals in terms of counts (note that an indicator variable is assumed to be equal to 1 unless explicitly stated):}
         \hat{Pr}(e_{t, d_{r}, c_{r}}|e_{t, d_{r-1}, c_{r-1}} \cap e_{t, d_{r-2}, c_{r-2}}) &= \frac{n(e_{t, d_{r}, c_{r}} \cap e_{t, d_{r-1}, c_{r-1}} \cap e_{t, d_{r-2}, c_{r-2}})}{n(e_{t, d_{r-1}, c_{r-1}}\cap e_{t, d_{r-2}, c_{r-2}})}  \\
	\intertext{We can now estimate the probability of a spike in the neurons' preferred channel $g$, in the time period $(t,t+\delta t]$, or $g_{t} = 1$, given an arbitrary history sequence:}
	\hat{Pr}(g_{t}| e_{t, d_{0}, c_{0}} \cap e_{t, d_{1}, c_{1}} \cap e_{t, d_{2}, c_{2}}) &= \frac{n(g_{t} \cap  e_{t, d_{2}, c_{2}} \cap  e_{t, d_{1}, c_{1}} \cap e_{t, d_{0}, c_{0}})}{n(e_{t, d_{2}, c_{2}} \cap  e_{t, d_{1}, c_{1}} \cap e_{t, d_{0}, c_{0}})}
\end{align*}

This leads to the general formula to estimate the probability of $g_{t}=1$ given an arbitrary subsequence $s$ of length $L_{s}$:

\begin{equation}\label{EPSTConditional}
    \hat{Pr}(g_{t}| e_{t, d_{r}, c_{r}}, r = 0,1,...,L_{s} - 1) = \frac{n\left(g_{t}, \underset{r = 0,1,...,L_{s} - 1}{\cap}e_{t, d_{r}, c_{r}}\right)}{n\left(\underset{r = 0,1,...,L_{s} - 1}{\cap}e_{t, d_{r}, c_{r}}\right)}
\end{equation}

The counts $n\left(g_{t}, \underset{r = 0,1,...,L_{s} - 1}{\cap}e_{t, d_{r}, c_{r}}\right)$ and $n\left(\underset{r = 0,1,...,L_{s} - 1}{\cap}e_{t, d_{r}, c_{r}}\right)$ are stored within the tree that has a preference for the channel $g$ at the tree node addressed by the sequence $s = e_{t, d_{r}, c_{r}}=1$, $r \in 0,1,...,L_{s} - 1$.

To achieve a probability estimate $g_{t+n}$, we must observe that it is likely that many results similar to Equation \ref{EPSTConditional} can be calculated for every time step $t+n$, one for each subsequence identified within the history window. Furthermore, there is no guarantee that any of these calculations will agree on a single conditional probability estimate for $g_{t + n}$. In other words, it is not straightforward to create a joint conditional $\hat{Pr}(g_{t+n}|s_{i}, i = 0,...,N_{s})$ from the individual conditionals $\hat{Pr}(g_{t+n}|s_{i}),\quad i = 0,...,N_{s}$. This inconsistency is addressed in the following Section \ref{EPSTRepresentativePattern}.

\subsection{Selecting a Representative Subsequence via a Reliability Measure}
\label{EPSTRepresentativePattern}

The entropy of the conditional probability estimate $\hat{Pr}(g_{t + n}|s)$ provides a simple heuristic to measure how reliable a particular subsequence $s$ is for predicting the state of $g_{t + n}$:

\begin{equation}\label{EntropyEPST}
    H(\hat{Pr}(g_{t + n}|s)) = \sum_{g_{t + n}\in\{0,1\}}\hat{Pr}(g_{t + n}|s)\log\left(\hat{Pr}(g_{t + n}|s)\right)
\end{equation}

This quantity has the advantage that it only considers the variability of $g_{t + n}$ given the feature $s$ and not vice versa. Given $s=1$ the entropy is estimated using counts:

\begin{align}
    \hat{Pr}(g_{t + n}=0,s=1) &= \frac{n(s=1) - n(g_{t + n}=1,s=1)}{t}\nonumber\\
    \hat{Pr}(g_{t + n}=0|s=1) &= \frac{n(s=1) - n(g_{t + n}=1,s=1)}{n(s=1)}\nonumber\\
    \hat{Pr}(g_{t + n}=1|s=1) &= 1 - \hat{Pr}(g_{t + n}=0|s=1)\nonumber\\
    H(\hat{Pr}(g_{t + n}|s)) &= \hat{Pr}(g_{t + n}=0|s=1)\log\left(\hat{Pr}(g_{t + n}=0|s=1)\right) \nonumber\\ 
    &+  \hat{Pr}(g_{t + n}=1|s=1)\log\left(\hat{Pr}(g_{t + n}=1|s=1)\right) \label{EPSTEntropyCount}
\end{align}

\begin{figure}
	\centering
	\includegraphics[width=1.0\textwidth]{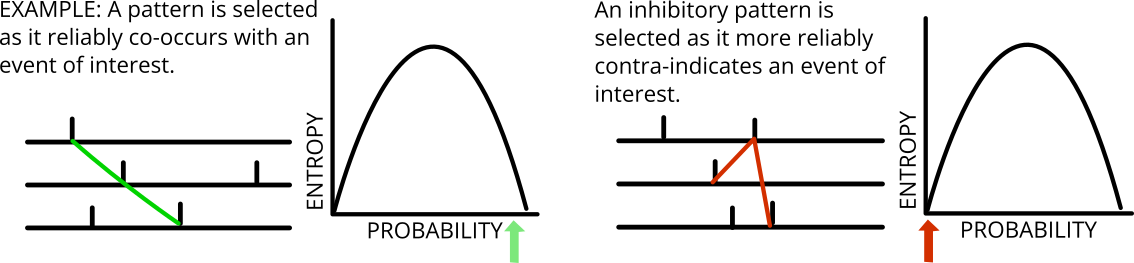}
	\caption{\label{EPSTEntropy} \textbf{A simple schematic diagram illustrating the entropy subsequence selection heuristic} The left panel shows a single subsequence selected for low entropy as a result of high 
 estimated probability $\hat{Pr}(g_{t + n}|s)$, this is the most frequent case for the normal action of the EPST. The right panel shows an interesting case where an additional spike matches a subsequence with lower entropy of the conditional, however, this time due to low estimated probability $\hat{Pr}(g_{t + n}|s)$. A deeper discussion of inhibitory patterns and how they may be used and learned is presented in Section \ref{EPSTXORProblem}.}.
\end{figure}

For the remainder of this paper, we use Equation \ref{EPSTEntropyCount} as the primary heuristic for selecting a subsequence to represent the history window $W_{t+n}$ given a channel $g$. The properties of the entropy measure are described in Figure \ref{EPSTEntropy}, and an additional computational benefit is discussed in Section \ref{EPSTXORProblem}.

\section{EPST Variants and Hyperparameters}

The EPST assumes the following:
\begin{enumerate}
    \item {There should be a finite, relatively small number of patterns generated by the dataset.}
    \item {The dataset should not change its statistics over time (should be ergodic).}
    \item {For each event there should be a single subsequence that can predict it without a contradicting pattern. If there exists a subsequence $s'$ such that $P(g_{n}|s') \neq 0$, it should be unlikely that there is another overlapping or non-overlapping pattern $s''$ for which $P(g_{n}|s'\cap s'') = 0$. }
    \item {Features should not be informative over long periods. For example, if a subsequence and a co-occurring event are separated by a delay longer than the history window, it will not be predicted by the EPST.}
\end{enumerate}

Assumption 4 is regularly violated by real life data sets. In traditional neural networks, the Long-Short Term Memory (LSTM) and other derivatives are used to combine longer term context with the short term (\cite{LSTMPaper}), however, there is no current mitigation available for the EPST and it is left to future work.

Real life data sets also do not often satisfy assumptions 1 and 2. This leads to the EPST taking on a large number of subsequences leading to an irreversible state for the EPST with many adverse performance effects. For this paper, we will name this issue the \emph{Tree Size Problem}, and potential solutions and mitigations for this problem are discussed in Section \ref{EPSTTreeSizeProblem}.

Assumption 3 can sometimes be broken as a side effect of the issues arising from the Tree Size Problem. However, some data sets may contain situations where either one of two subsequences predict an event, but not both. This is analogous to an XOR operation, and as such is referred to as the \emph{XOR Problem}, and is discussed in Section \ref{EPSTXORProblem}.

These situational limitations necessitate the development of EPST variants and additional functionality to handle the specific requirements of the application.

\subsection{Tree Size Problem}
\label{EPSTTreeSizeProblem}

Due to the combinatorial nature of the EPST subsequence collection method, the maximum tree depth (or sequence length) should be limited to avoid unnecessarily high memory usage. The EPST also requires a minimum subsequence length when making predictions. Using only medium length subsequences as features confers the following advantages:
\begin{itemize}
    \item {\textbf{Avoiding rare features} Storing long subsequences of precise timings would be wasteful, as overly long subsequences are unlikely to arise more than once in a dataset.}
    \item {\textbf{Avoiding common features} Storing short subsequences is unavoidable as they are subsequences of any longer sequences. Short subsequences however are not informative for use in predictions. This is because they are extremely common, and are frequently created by noise within a dataset, or interactions between independent patterns.}
    \item {\textbf{Designing for efficiency} The original PST \cite{vmmpst} defines mechanisms to limit the length of branches of the tree data structure. This is due to the fact that the total number of possible stored sequences increases exponentially with subsequence length. The EPST also limits tree depth by including a subsequence length limit hyperparameter and limits the speed of tree growth.}
\end{itemize}

 By defining a minimum length hyperparameter, and allowing the EPST to only make inferences using sequences longer than this, we can begin to mitigate against random erroneous clashes between the EPST data structure and the dataset. The second mitigation is the inclusion of an additional parameter limiting the maximum depth the tree is allowed to grow. The \emph{frequency threshold} hyperparameter is inherited from the PST and only allows a subsequence to be used in a probability estimate if the denominator count is greater than the hyperparameter amount.

Rapid tree growth may also be attributed to rare uninformative interactions between independent patterns or noise patterns. The two hyperparameters \emph{maximum spike interval} and \emph{branch extension threshold} are designed to mitigate against rapid tree growth. The \emph{branch extension threshold} only allows a node to be added to the tree if the numerator count of the node is greater than the hyperparameter amount. This is a learning rate parameter intended to trade off the learning speed against the reliability of patterns stored in the tree. The \emph{maximum spike interval} hyperparameter on the other hand does not store subsequences if the delay between subsequent events is greater than the hyperparameter amount. Using this hyperparameter expresses the assumption that events are less dependent on one another with decreased proximity in time.

If both the \emph{branch extension threshold} and the \emph{frequency threshold} parameters are set to zero, the EPST is capable of learning sequences after a single presentation.

In addition to these hyperparameters, pruning may also be employed to reduce the number of stored patterns within the EPST data structure. Pruning is executed at a user defined interval. The procedure is also provided with a constant threshold for the reliability of the pattern (Equation \ref{EPSTEntropyCount}), all patterns measured above this threshold are removed. It is also possible to prune stored subsequences uniformly randomly sampled from the EPST model. This is hypothesised to cause minimal degradation in prediction performance due to high pattern redundancy (Figure \ref{EPSTtoPST3}), and also due to sequential sampling as outlined in Section \ref{EPSTEfficiency}.

\subsection{XOR Problem}
\label{EPSTXORProblem}

\begin{figure}
	\centering
	\includegraphics[width=1.0\textwidth]{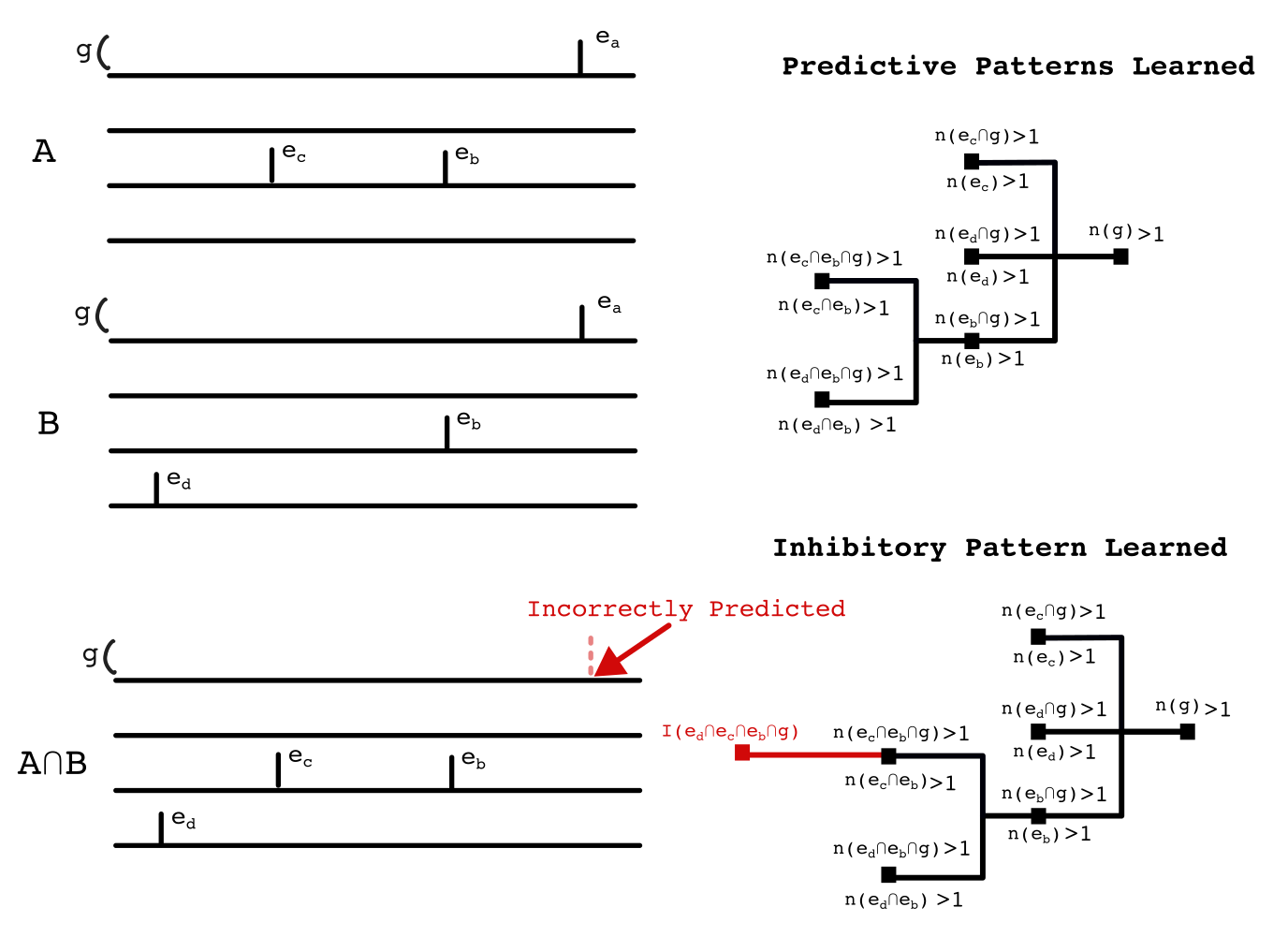}
	\caption{\label{EPSTInhibition} \textbf{An illustration of a new inhibition system to extend the EPST.} The top panel shows 2 separate occasions where a simple subsequence is learned to indicate an event $e_{a}$. At this point, all counts across all nodes in the tree hold counts $>1$. Any subsequences matched which satisfy the thresholds outlined in Section \ref{EPSTTreeSizeProblem} will put forward a non-zero probability estimate, or $\hat{Pr}(g_{t + n}|s) > 0$ for all subsequences $s$ stored in the tree. This presents a problem as the lower panel shows a later instance where both patterns combined indicate a false positive $e_{a}$. As a solution, the inhibitory pattern is formed by the addition of a single inhibitory node. If this node is matched in the future, it will contribute an estimate of zero to the set of matches $\hat{Pr}(g_{t + n}|s_{I}) := 0$, which will likely be chosen by the entropy heuristic discussed in Section \ref{EPSTRepresentativePattern}.}
\end{figure}

The EPST as defined in Section \ref{implementationdetails} is unable to implement an XOR operation between two subsequences. This issue is illustrated in Figure \ref{EPSTInhibition}, along with an extension to the EPST algorithm that provides a solution. The figure shows that pattern A ($e_{c}\cap e_{b}$) and B ($e_{d}\cap e_{b}$) can independently be learned as indicators of a spike at $e_{a}$, and both $A\cap B$ together ($e_{d}\cap e_{c}\cap e_{b}$) may subsequently be learned as an indicator of no spike at $e_{a}$ via inhibition. This is made possible by keeping track of incorrect predictions made by excitatory subsequences such as A and B. When a mistake is made, all subsequences from the window are collected that \emph{do not currently exist} as excitatory patterns, and these patterns are stored as inhibitory subsequences. This additional measure of not adding inhibitory subsequences that already exist as excitatory sequences protects the mechanism from event noise such as dropout.

Jitter on the other hand is more complex still, as it typically takes a long time to collect all relevant excitatory patterns (Section \ref{EPSTJitterSection}). This requires extra mitigations, such as allowing a stored inhibitory subsequence to be destroyed if it causes false negatives.

Inhibitory subsequences may be stored in the EPST as normal patterns in the tree with a numerator initialised to count 0 and a denominator initialised to count 1 (Equation \ref{SimpleConditional}). However, this may add an unacceptable number of patterns to the tree exacerbating issues discussed in Section \ref{EPSTTreeSizeProblem}. A more robust implementation explicitly stores inhibitory patterns with a single joint count $n(e_{d}\cap e_{c}\cap e_{b}\cap g)$. If the count exceeds a globally defined threshold the pattern is pruned. Regardless of how an inhibitory pattern is stored, if the pattern is matched from the window $W_{t + n}$ during prediction, the probability estimate is assigned a minimum value $\hat{Pr}(g_{t + n}|s_{I}) := 0$ where $s_{I}$ is the inhibitory subsequence.

\section{Comparison with Traditional VMMs Under Common Event Noise Conditions}
\label{VMMComparisonChapter}

In this section, we will compare performances of the EPST to existing VMM algorithms, the PPM-C and PST. These algorithms were chosen as the EPST shares many properties with the VMM, however, there may be challenges in comparison due to the differences between event based and sequential symbolic data. It is for this reason that the dataset and comparison methods must be carefully designed.

\begin{figure}
	\centering
	\includegraphics[width=1.0\textwidth]{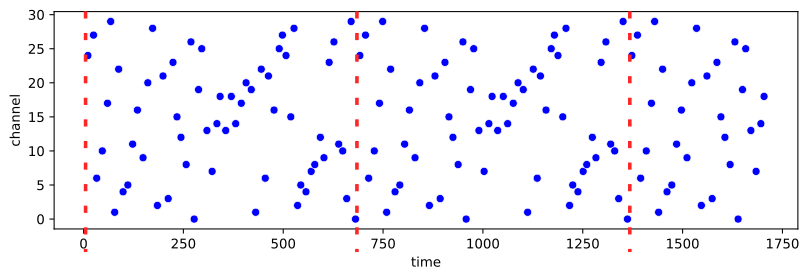}
	\caption{\label{VMMComparisonData} \textbf{Example dataset used for the VMM vs EPST comparison.} The data is generated pseudo-randomly based on a seed for each presentation. The delay between events is an integer between 8 and 14 time steps, and events are distributed evenly across 30 channels. In order to provide sequences to learn, the data is cyclic and repeats every 60 events; dashed red lines show the boundaries of this cyclic period.}
\end{figure}

We designed an artificial dataset that is simple enough to illustrate the features of the event based algorithms as compared to the VMM algorithms. The base dataset is designed to be easy for both the sequential algorithms and the EPST to learn online and follow, such that the effects of the event noise scenarios presented in the following sections are easy to observe and attribute. Each scenario represents a common form of event noise, such as additive events or removing events.  We will use a simple base dataset of a precisely repeating cyclic pattern across 30 channels, an excerpt is shown in Figure \ref{VMMComparisonData}. For each experiment in this chapter, 25 different base datasets were applied and the errors averaged for each results plot. 

Aside from the standard EPST, three variants of the EPST will also be compared for every experiment in this chapter. The variants considered are the EPST with inhibition (EPST\_I), the EPST with pruning (EPST\_P), and the EPST with both inhibition and pruning (EPST\_IP). The inhibition extension is described in Section \ref{EPSTXORProblem}. The pruning extension is a simple procedure which removes any sequences from the tree that have a calculated entropy (Section \ref{EPSTRepresentativePattern}) greater than 0. This procedure executes every 500 events received by the EPST and is intended to clear any potentially unreliable patterns from the tree.

In the following sections, we will investigate the effects of additive structured noise, additive random noise, event time jitter, and event time jitter with dropout. These scenarios were chosen to best illustrate the strengths, and some weaknesses, of the novel EPST algorithm. Finally, in Section \ref{ChallengeComparisonToVMM}, we discuss the challenges of comparing the prediction performance of the EPST to sequential VMM algorithms, and illustrate the solutions engineered to overcome these challenges.

\subsection{Predicting Two Superimposed Structured Sequences}
\label{EPSTStructuredInterferenceSection}

\begin{figure}
	\centering
	\includegraphics[width=1.0\textwidth]{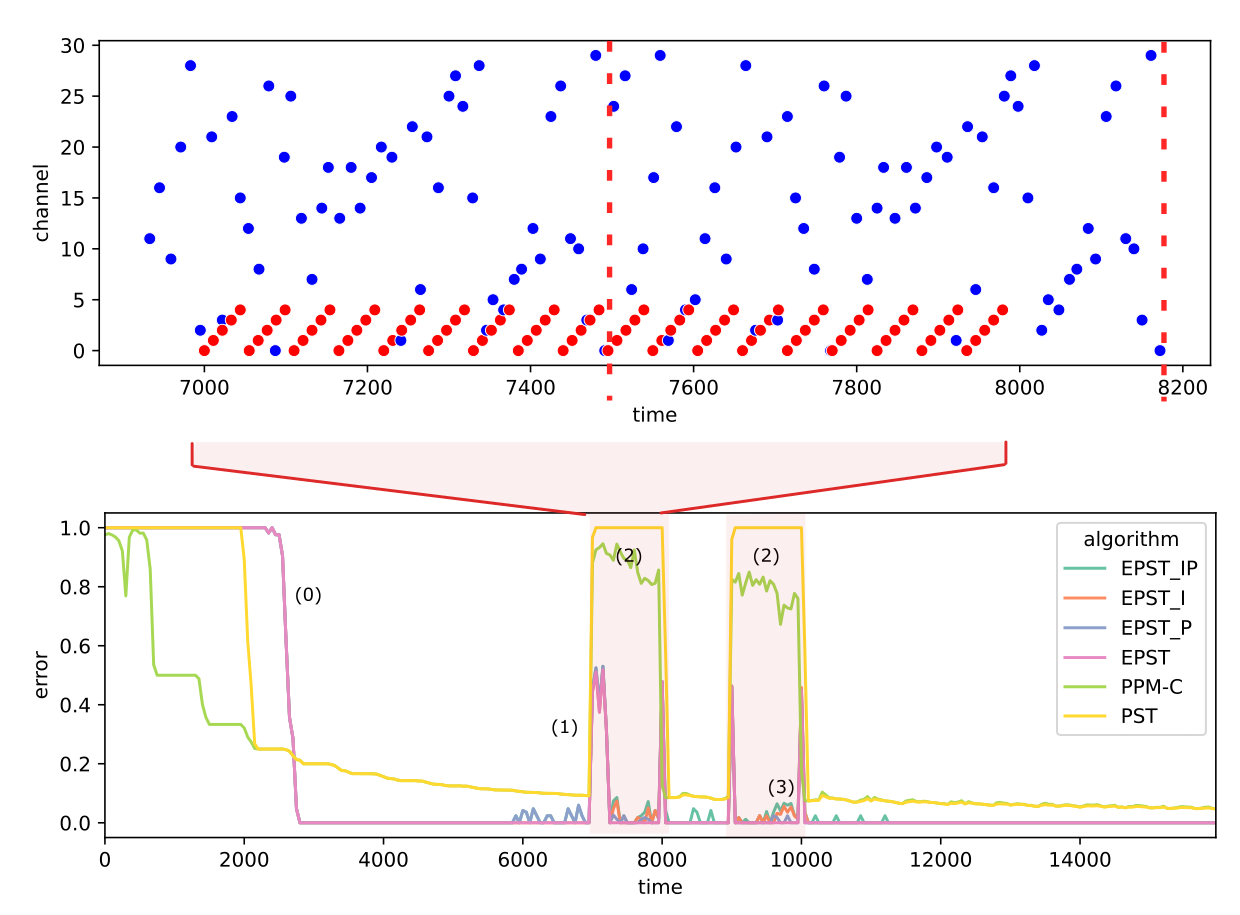}
	\caption{\label{ComparisonPlot} \textbf{Results plot for the structured interference experiment using the same pattern for each interference region}
  The lower panel shows an overview trace of an experiment run. The time step number is given on the x-axis, and the y-axis shows the average difference between the true probability and the calculated probability (This is an estimated error, and the method for calculating this is discussed in Section \ref{ChallengeComparisonToVMM}). The region labelled (0) shows decreasing error during learning. The label (1) refers to a brief period of increased error where the EPST is learning the newly applied structured interference. The red regions labelled (2) illustrate where structured noise was applied and the top panel shows an example of the structured interference (red dots) laid over the signal (blue dots). A discussion of the slight bump in error of the EPST labelled by (3) is deferred to Section \ref{EPSTRandomClashes}. }
\end{figure}

 The first experiment tests the robustness of all algorithms to a simple repeating signal overlaid on top of the existing test signal. A mockup of this interference is shown at the top of Figure \ref{ComparisonPlot}. The challenge is for algorithms to identify and learn all patterns in parallel with no prior knowledge. For this experiment, the EPST has a history window of 32 time steps and only assigns probabilities to history subsequences of 3 or more events. As for the VMMs, the PST and PPM-C were assigned a sequence length of 8, and the only other parameter of note for online learning for the PST is the minimum frequency required to make a probability calculation which was set to 3.

We expect the relative behaviour and accuracy of all three algorithms to be similar over time under normal circumstances using the simple dataset, and this is evident from the trace in Figure \ref{ComparisonPlot} between time steps 3000-7000. The first difference between the EPST and the sequential VMMs is the learning period between time steps 0 and 3000, label (0). The length of this learning period for the EPST is determined by the branch extension threshold, which was set to 1 (Section \ref{EPSTTreeSizeProblem}). The higher the EPST branch extension threshold parameter, the longer the learning period. Interference was applied between times 5000-6000 and 7000-8000. After a short period of learning the interference in the first interval, all EPST algorithms perform well. The traditional algorithms, however, without the ability to use timing information, fail to follow both patterns simultaneously even after many presentations. 

The high performance of the EPST during the interference is an example of the `Multiple trajectory tracking' property, where the EPST can follow several sequences in parallel up to some probabilistic capacity depending on the dataset and parameterisation.

\begin{figure}
	\centering
	\includegraphics[width=1.0\textwidth]{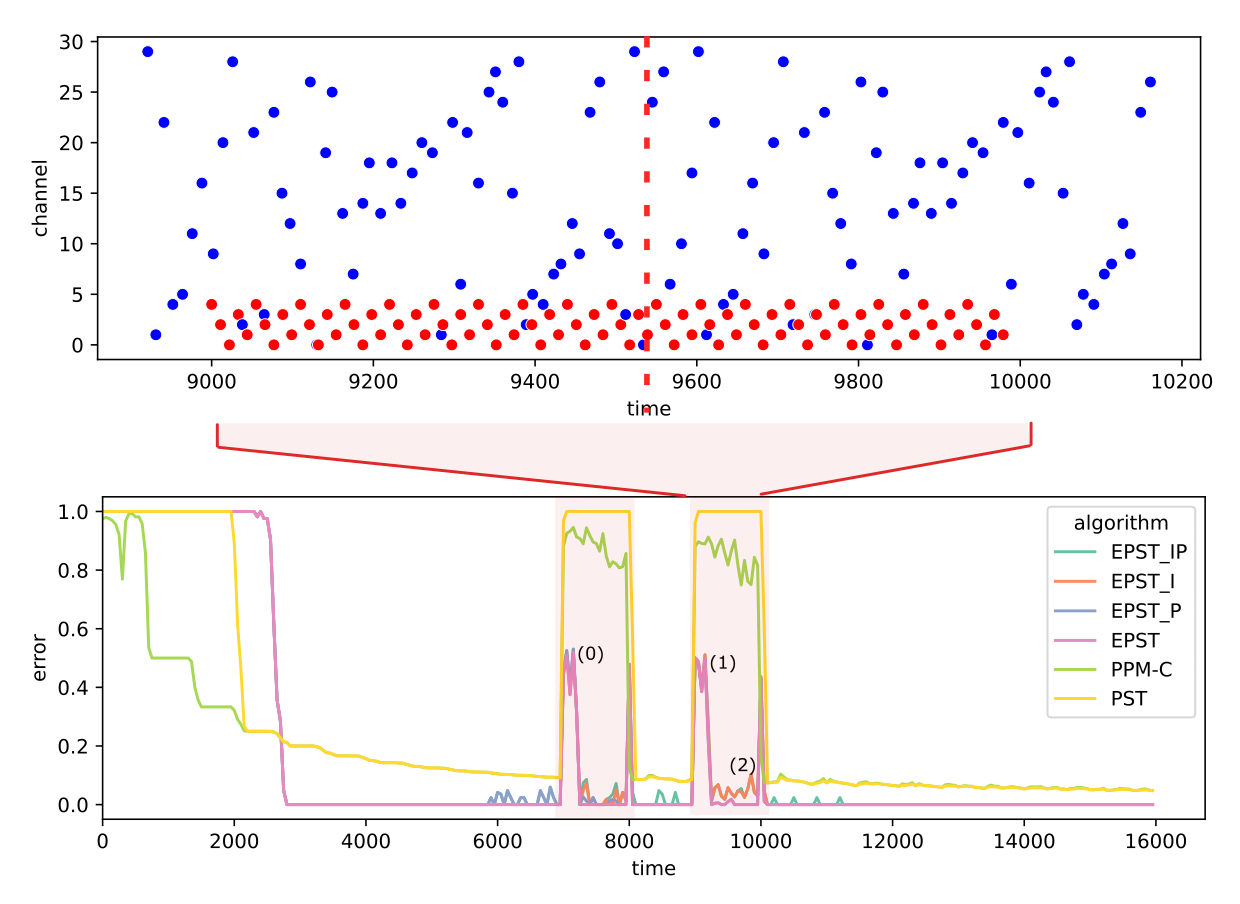}
	\caption{\label{ComparisonPlotAlt} \textbf{Results plot for the structured interference experiment using a different pattern for each interference region} The label (0) indicates the first interference region, which uses the same interference pattern as illustrated in Figure \ref{ComparisonPlot}. The label (1) indicates a brief period of high error for the EPST as it learns the new interference pattern as it is different to the first. The second interference pattern is illustrated in the top panel. Discussion of the slight bump in error of the EPST labelled by (2) is deferred to Section \ref{EPSTRandomClashes}. }
\end{figure}

Figure \ref{ComparisonPlotAlt} shows a slight variant of the structured interference experiment in Figure \ref{ComparisonPlot} where the second interference region is replaced by a different interference pattern. The main difference between experiments is that the second interference pattern in Figure \ref{ComparisonPlotAlt} shows a similar bump in error labelled (1), where the second interference pattern is learned by the EPST.

One feature of particular note is the high reported error of the PST under noise conditions. This is due to the definition of the algorithm provided in \cite{vmmpst} not having a mechanism to assign a probability to sequences with a low number of observed instances, which defaults to a maximum error in the case where no probability is calculated.

\subsection{Random Clashes with Stored Patterns}
\label{EPSTRandomClashes}

\begin{figure}
	\centering
	\includegraphics[width=1.0\textwidth]{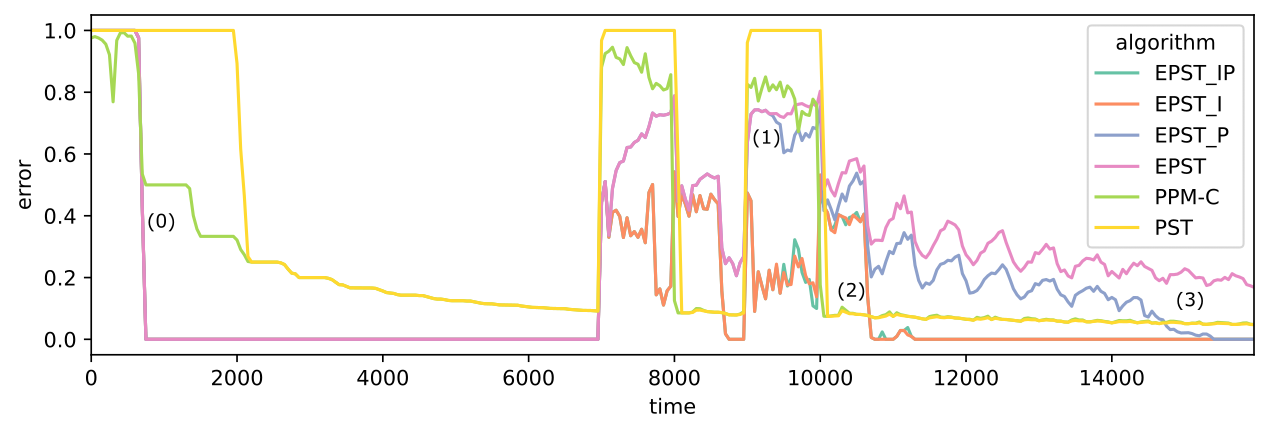}
	\caption{\label{ComparisonPlotET0} \textbf{Results plot for the structured interference experiment using the same pattern for each interference region, using branch extension threshold of 0.} A diagram showing results of a similar experiment to Figure \ref{ComparisonPlot}, with the branch extension threshold parameter set to 0. Label (0) refers to the relatively short learning period. Label (1) (EPST\_P) shows the small and brief effect that pruning has on the error rate during interference. Label (2) (EPST\_I, EPST\_IP) shows that the inhibition extension can allow a quick reduction in error rate following the end of the interference. Label (3) shows the long tail of the error rate for both the raw EPST and EPST\_P as the node probability estimates are improved over time by counts. Only EPST\_P can achieve an error rate of 0 through the removal of unhelpful subsequences.}
\end{figure}

The EPST shows a subtle decay in performance towards the end of the experiment in Figure \ref{ComparisonPlot} label (3) and \ref{ComparisonPlotAlt} label (2). This issue is illustrated to the extreme in Figure \ref{ComparisonPlotET0}, where the EPST branch extension threshold is set to its minimum value of 0, all else equal.

The branch extension threshold of 0 allows every subsequence from the history window to be added to the tree, regardless of the number of times it is encountered. On one hand, this can lead to one-shot learning as illustrated in Figure \ref{ComparisonPlotET0} label (0), where only one cycle of the data is required to learn the signal. However, the downside of this approach is a rapid tree growth, leading to issues associated with large tree sizes (Section \ref{EPSTTreeSizeProblem}), illustrated by the poor performance beyond timestep 7000.

Despite actually increasing the number of stored subsequences in each EPST unit, the inhibition extension (EPST\_I, EPST\_IP) was extremely effective at reducing the error rate. Error rates decrease after the onset of the interference and are eliminated shortly after the interference is removed, as illustrated at label (2). The pruning routine chosen (EPST\_P, EPST\_IP) is not shown to be as effective at reducing the error rate. This is hypothesised to be because pruning does not stop subsequences from being re-learned after they are removed. To support this hypothesis, label (1) shows a brief reduction in error rate caused by pruning which quickly returns. This means that the recovery to a pre-interference error rate can only begin after interference has stopped, as illustrated by the label (3). 

\begin{figure}
	\centering
	\includegraphics[width=0.9\textwidth]{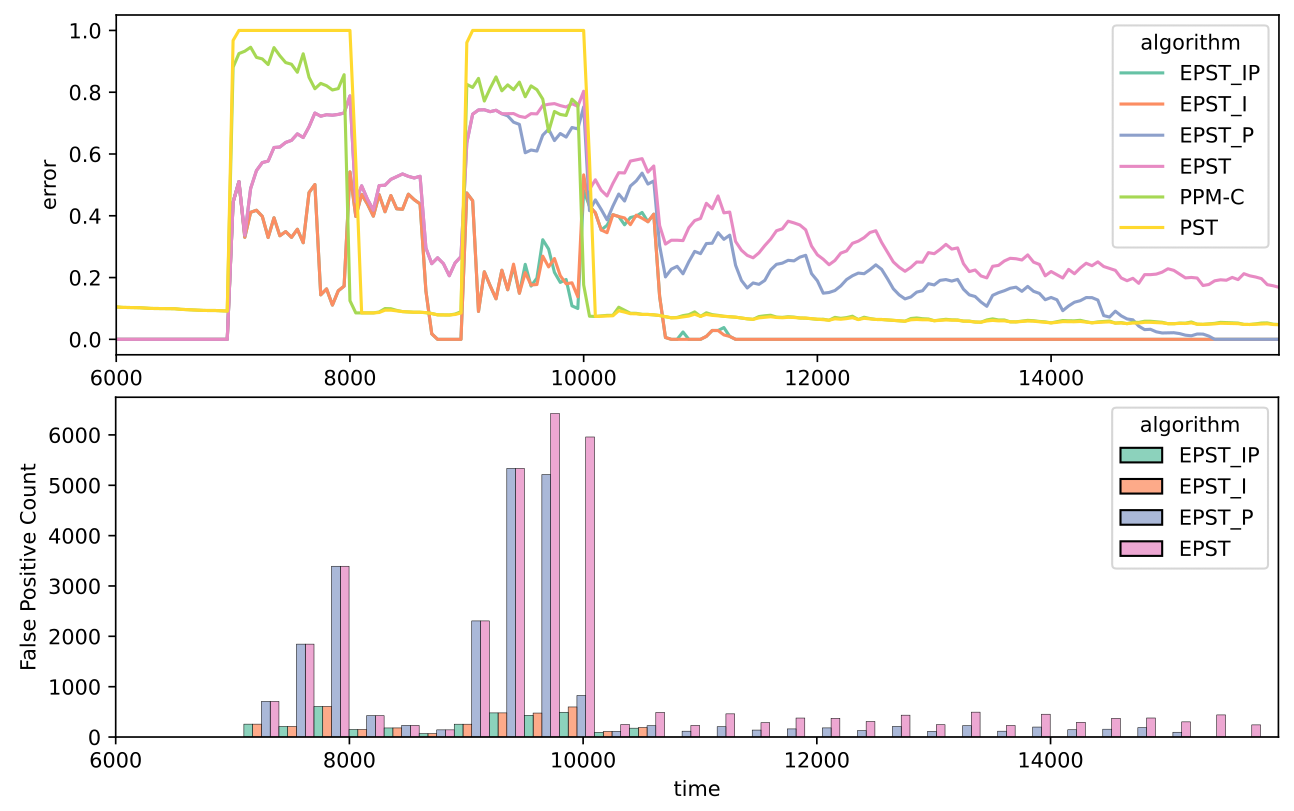}
	\caption{\label{ComparisonPlotET0FP} \textbf{A detailed analysis of the false positives associated with the experiment in Figure \ref{ComparisonPlotET0}.} The top panel shows an excerpt of results from Figure \ref{ComparisonPlotET0}. The lower panel shows the binned false positive counts for all EPST algorithms observed over the same region of time.}
\end{figure}

The poor performance of the EPST in this experiment can be attributed to an increasing number of false positives, as shown by the histogram in Figure \ref{ComparisonPlotET0FP}. False positive counts on the y-axis explode during the periods of interference for the vanilla EPST. This is due to a problem faced by the EPST where the number of stored patterns becomes sufficient for random clashes with the test data to occur, resulting in false positives, as discussed in Section \ref{EPSTTreeSizeProblem}. The plot also shows that this problem is mitigated somewhat by the EPST variants. The pruning extension alone (EPST\_P) reduces false positives to a fraction of their peak at time step 10000, but only eliminates the majority of false positives by time step 15000. The variants that include inhibition  (EPST\_I, EPST\_IP) on the other hand eliminate all false positives by time step 11000. The inhibition extension achieves a false positive reduction by actively adding and retaining additional patterns that prevent false positives. 

\subsection{Predicting Structured Sequences in the Presence of Added Random Events}
\label{EPSTRandomInterferenceSection}

\begin{figure}
	\centering
	\includegraphics[width=1.0\textwidth]{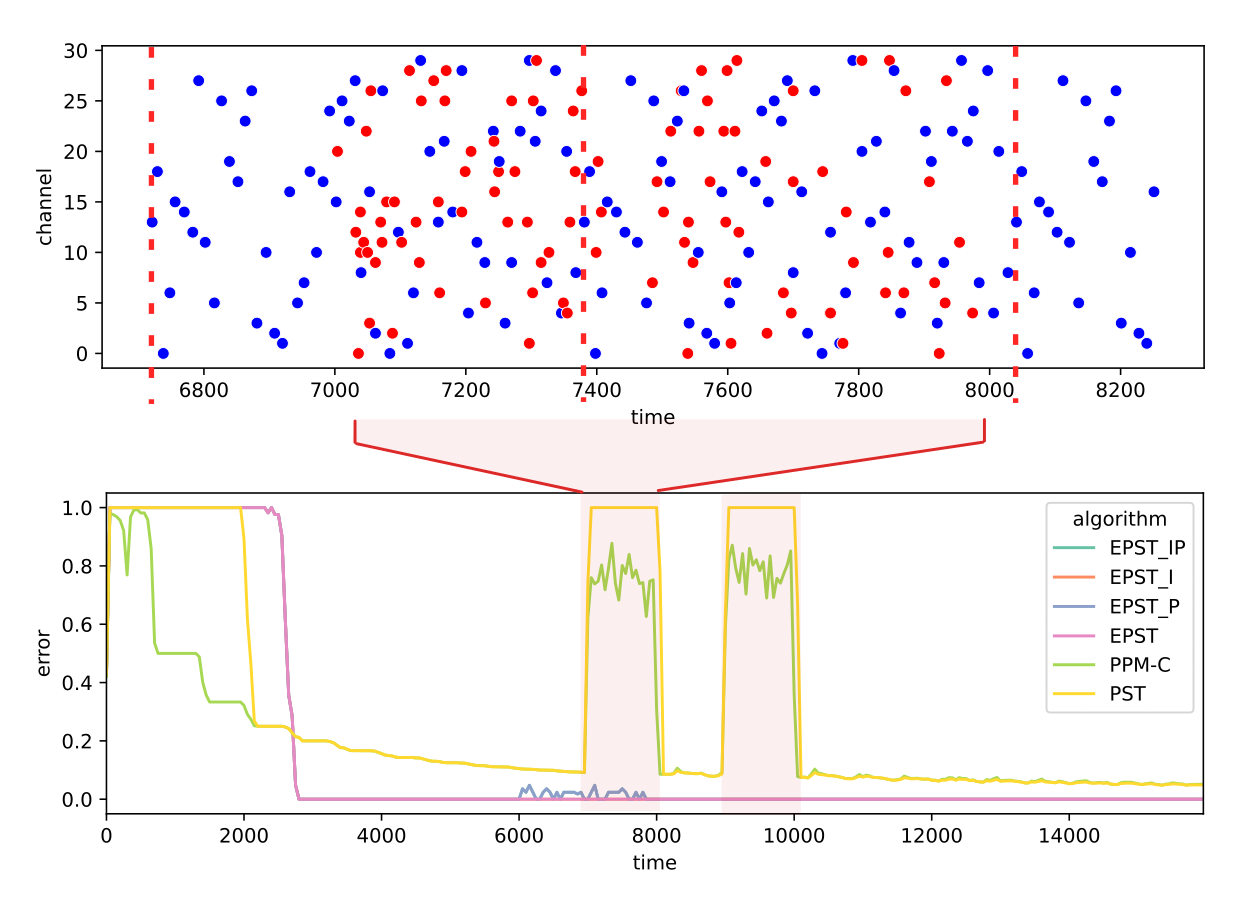}
	\caption{\label{ComparisonAddRandomNoise} \textbf{Results plot for the random interference experiment.} 
        The top panel shows an example of the randomly sampled interference (red dots) laid over the signal (blue dots). The lower panel shows the results of the random additive event noise experiment. Along the x-axis is time and along the y-axis is average error. The intervals where interference is applied are indicated by red shading.
 }
\end{figure}

The random event addition experiment illustrated in Figure \ref{ComparisonAddRandomNoise}, while similar to the interference experiment shown in Figure \ref{ComparisonPlot}, differs in that the inserted interference events have no predictable pattern and cannot be learned over time. Due to the fact that independently sampled noise events cannot reasonably be predicted, the error calculation method is changed to ignore interference events, and to instead focus on how the noise affects the accuracy of the prediction of the signal (See Section \ref{ChallengeComparisonToVMM}).

The top panel of Figure \ref{ComparisonAddRandomNoise} shows an example excerpt of the data with the interference applied. The interference itself consists of 100 events, sampled uniformly across 1000 time steps and all 30 channels. Interference was applied between times 7000-8000 and 9000-10000, this is illustrated in the results plot in lower panel of Figure \ref{ComparisonAddRandomNoise}. The EPSTs ability to predict the signal is completely unaffected by the random noise, and no increase in error is observed. The VMM algorithms, however, without the ability to use timing information, fail to follow the signal while the random interference is applied.

One interesting observation is that EPST appears to perform better in this random event interference experiment than its structured analogue. This is hypothesised to be due to the interactions between the cyclic signal and structured interference creating subsequences that are repeated enough to be stored within the EPST data structure. These subsequences may not be informative, and can cause false positives that slightly increase the error rate of the EPST.

\subsection{Predicting Structured Sequences With Random Jitter in Event Times and Randomly Dropped Events}
\label{EPSTJitterSection}

\begin{figure}
	\centering
	\includegraphics[width=0.9\textwidth]{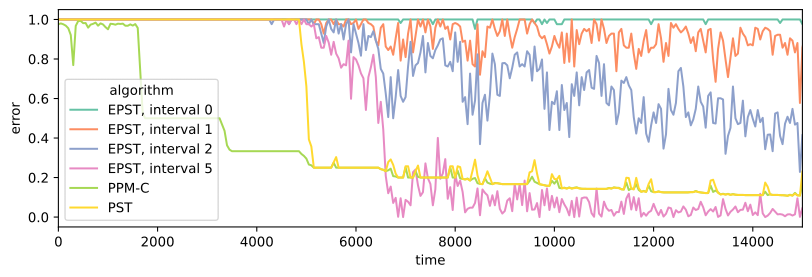}
	\caption{\label{ComparisonAddJitter} \textbf{Results plot for the random time jitter experiment.} Along the x-axis is time and the y-axis is average error. Jitter values from the interval $[-4, 4]$ are applied after event 500 ($\approx$ 5500 time steps) for the remainder of the scenario.}
\end{figure}

Jitter, a form of event noise, occurs when event times are offset from an assumed true time point by a random value. For the scenario shown in Figure \ref{ComparisonAddJitter}, the random jitter is an integer sampled uniformly from the interval $[-4, 4]$ measured in time steps. For the jitter experiments, the EPST variants compared differ only in the `matching interval size' parameter. The matching interval controls the width of a time interval surrounding a stored spike for which an input spike can be considered equal (Figure \ref{time_matching_methods}).

Figure \ref{ComparisonAddJitter} shows that the VMMs handle event jitter very well. This makes sense as VMMs are order based, and as long as event jitter does not change the order of events the representation is not affected. The EPST on the other hand performs extremely poorly at matching interval size 0, which corresponds to a matching interval of a single time bin. This is because jitter from the interval $[-4, 4]$ creates too many combinations of 3-length subsequences to be learned within the experiment time frame. Increasing the matching interval size allows the EPST to perform better, with performance equalling or exceeding that of the order-based VMMs at matching interval 5.

\begin{figure}
	\centering
	\includegraphics[width=0.9\textwidth]{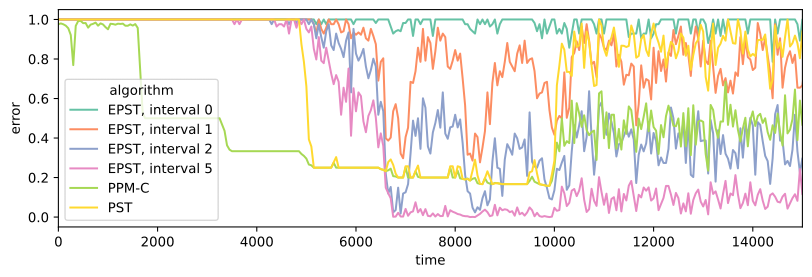}
	\caption{\label{ComparisonAddJitterDropout} \textbf{Results plot for the random time jitter experiment including the application of random dropout.} Along the x-axis is time and the y-axis is average error. Jitter values from the interval $[-4, 4]$ are applied after event 500 ($\approx$ 5500 time steps) for the remainder of the scenario. 20\% of events were removed randomly after simulation time 10000, leading to a visible increase in error rate for all algorithms, especially the PST and PPM-C.}
\end{figure}

 To illustrate why one might choose the EPST over the traditional VMMs even under event jitter conditions, we will design an additional experiment with both time jitter and dropout noise types. Dropout in this case refers to an event that randomly fails to transmit entirely. The scenario results are shown in Figure \ref{ComparisonAddJitterDropout}, where dropout begins at time 10000 and affects 20\% of events. 

The plot demonstrates that the EPST remains robust to dropout even under jitter conditions, as long as the equivalence interval hyperparameter is sufficiently large. The VMMs perform poorly after the application of dropout, as they lack a robust mechanism to handle missing data. 

Additionally, it was confirmed that the inhibition mechanism was robust to both jitter and dropout, as anticipated from Section \ref{EPSTXORProblem}. The plots generated for the inhibition variant appear identically to Figures \ref{ComparisonAddJitter} and \ref{ComparisonAddJitterDropout}.

\subsection{Quantifying the Prediction Performance of the EPST to Allow Comparison to the VMM}
\label{ChallengeComparisonToVMM}

Performance in the following experiments is measured as an average probability error over time, where time is measured in time bins. It is simple to see how this performance measure can be calculated for a sequential VMM, by averaging probability errors from each event over a predefined time window. However, this presents a challenge to map the probability estimates of the EPST onto the same estimate as the VMMs. 

The term $\hat{Pr}_{EPST}(c|W_{n})$ can be read colloquially as the estimated probability calculated by the EPST of an event in channel $c$ at time step $n$. Where a time step is described by $t$ such that a time step is defined $(t, t+\delta t]$, and indexed by $n$. On the other hand, the VMM simply outputs the estimated probability of the channel of the next event, regardless of when this event takes place. Upon receiving an event at time $t$ in channel $c$, we may marginalise the probability estimates from the EPST in channel $c$ over time to find a comparable output value to the VMM. The marginalised probability is normalised by the sum of all marginal probabilities across all channels. This formula should allow the probabilities reported by the EPST to be compared to those output by the VMMs:
\begin{equation} \label{EPSTvsVMMEvalEq}
    \hat{Pr}_{VMM}(\text{Next event is in }c) \approx \frac{\sum_{t'=t_{prev}...t_{event}}\hat{Pr}_{EPST}(c|W_{t'})}{\sum_{\forall c'}\sum_{t'=t_{prev}...t_{event}}\hat{Pr}_{EPST}(c'|W_{t'})}
\end{equation}

In addition to the problems addressed above, the noise scenarios contain nonlinear and overlapping streams of events. This presents issues for the evaluation of the EPST as correct estimates may be penalised by Equation \ref{EPSTvsVMMEvalEq} incorrectly in common circumstances. Since we have access to the full dataset generating the spike sequence in advance (Figure \ref{VMMComparisonData}), we know at evaluation time exactly where spikes from both signal and noise are. For the EPST this allows us to isolate the exact set of channels and times that should contain a non-zero prediction, so that we can identify in various situations (such as under interference conditions) which predictions should be included in the scoring and which should not.

\begin{figure}
	\centering
	\includegraphics[width=1.0\textwidth]{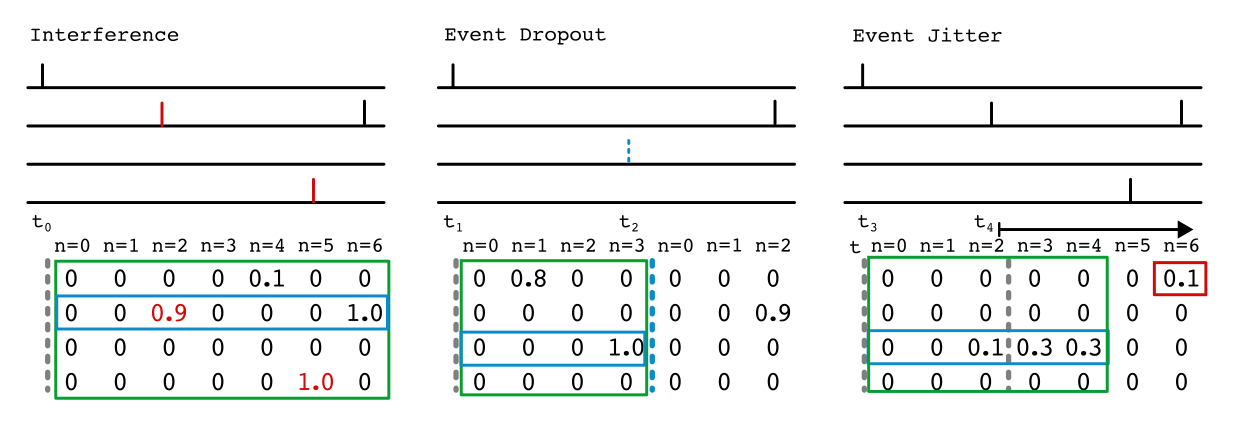}
\caption{\label{EPSTComparisonMethods} \textbf{A schematic showing a comparison of the different methods used in specific scenarios to calculate estimated probabilities and errors for the EPST.} As the sequential data predicted by the VMM is very different qualitatively to that handled by the EPST, we must engineer a way to compare their results. The output of a typical EPST spike triggered prediction at time $t$ is (from Section \ref{MathsProblemDefinition}) a $C\times M'$ matrix of probability estimates. This matrix is shown on the lower half of each of the three columns, one channel per row, and the corresponding true spikes from the data are presented at the top of the figure. Each column of the matrix represents the prediction at a time step $n$ of the simulation. The general process of calculating a single next spike probability for comparison to the VMM algorithms is to sum the channel row in the blue box and normalise it by the sum in the green box (Equation \ref{EPSTvsVMMEvalEq}). The details however differ for each of the experiments, and are described in Section \ref{ChallengeComparisonToVMM}.
 }
\end{figure}

As each noise scenario in this chapter presents a new challenge to compare event based and sequential algorithms, each case must be handled differently. An overview of these comparison methods is given in Figure \ref{EPSTComparisonMethods}. The left panel corresponds to structured interference (Section \ref{EPSTStructuredInterferenceSection}). The middle and right panel handles event dropout and jitter respectively, which are combined for application in Section \ref{EPSTJitterSection}. Note that crucially these adjustments are made during scoring, and no changes are made to the input for either of the EPST and VMM algorithms.

The left hand panel in Figure \ref{EPSTComparisonMethods} represents an example excerpt of the structured interference experiment (Section \ref{EPSTStructuredInterferenceSection}). The black events represent the signal, and red events represent the interference. In this scenario we are using an EPST spike triggered simulation beginning at $t_{prev}:=t_{0}$ to predict the next signal spike in the region $(t_{0}+6,t_{0}+6+\delta t]$. The individual time step probability estimates for the interference spikes in red, although correct, hinder our calculation so must be zeroed out before the EPST next spike formula is applied (Equation \ref{EPSTvsVMMEvalEq}). Conversely, predictions corresponding to the main signal time steps are zeroed to calculate the error for the interference sequence, while $t_{prev}$ is assigned to the most recent interference event.

In random interference (Section \ref{EPSTRandomInterferenceSection}), no attempt is made to predict the random noise, and only the signal is used in the error calculation. This amounts functionally to ignoring the error evaluation step for all random interference events. This means that for Equation \ref{EPSTvsVMMEvalEq}, $t_{prev}$ is always set to the most recent signal event, regardless of whether a subsequent interference event was received.

The middle panel in Figure \ref{EPSTComparisonMethods} represents an example excerpt of the event dropout experiment (Section \ref{EPSTJitterSection}). Events dropped on input to the EPST algorithm are treated as if they were not dropped for the purposes of calculating the error. This means that if an algorithm predicts a dropped event it is not incorrectly penalised. Again, the black events represent the signal. In this scenario we are using a spike triggered simulation at $t_{prev}:=t_{1}$ to predict the next signal spike which was dropped in the region $(t_{1}+3,t_{1}+3+\delta t]$ (blue dashed line). Following this error calculation, this dropped event time will now be assigned as $t_{prev}$ for subsequent calculations.

The right panel in Figure \ref{EPSTComparisonMethods} represents an example excerpt of the event jitter experiment (also Section \ref{EPSTJitterSection}). The upper bound on the summation in Equation \ref{EPSTvsVMMEvalEq} must be extended by a constant amount to cover the distribution of potential jitter values predicted by the EPST. In this final scenario we illustrate a spike triggered simulation beginning at $t_{prev}:=t_{3}$ to predict the next signal spike which arrives in $(t_{4},t_{4}+\delta t]$, and so the calculation window is padded to $t_{4}+3$. For the subsequent error calculation at the event in the time step $(t_{4} + 4,t_{4} + 4 +\delta t]$, we must assign $t_{prev}:=t_{4}$ and exclude all of the predictions used for prior events by zeroing out the numbers within the blue box.

\section{Discussion}

In this paper, we introduce the EPST, a novel event based prediction algorithm adapted from the PST. The EPST has been shown to outperform some commonly used sequential VMM algorithms, including the PST itself, under various event noise conditions. The event noise conditions explored in this chapter are: structured additive interference (Section \ref{EPSTStructuredInterferenceSection}), random additive interference (Section \ref{EPSTRandomInterferenceSection}), random event time jitter, and random dropout (Section \ref{EPSTJitterSection}). In the experiments, the EPST demonstrated robustness to structured additive noise, random additive noise, and dropout, experiencing a negligible reduction in accuracy while interference was applied. However, the event time jitter scenario is a challenge for the EPST. In this case, increasing parameters such as the matching interval (Figure \ref{time_matching_methods}) is shown to mitigate the issue while retaining the advantages of the EPST in other aspects.

Two key phenomena were identified as the main drivers of the superior performance of the EPST over its sequential symbolic counterparts:
\begin{itemize}
    \item {\underline{Multiple trajectory tracking} The EPST was found to be able to follow and predict multiple independent overlapping patterns at the same time. In practical examples of cyber security monitoring, as well as complex sensor data such as audio, many objects or trajectories of information may be present and superimposed into a single input stream of information. These kinds of superimposed input streams may be handled efficiently in an event based context, in a way that is not possible or intractable using sequence based data formats.}
    \item{\underline{Resistance to noise} Closely related to the multiple trajectory tracking feature of the EPST is the resistance to additive and subtractive noise. Noise events can be inserted into sequences that change the signal entirely if it is interpreted as a symbolic sequence such as in a typical VMM. Symbols can also fail to transmit or be detected, which also fundamentally changes the signal if interpreted as a symbolic sequence. For event based data, neither of these noise types change the timing relationships between all other non-noise events, allowing the EPST to maintain performance where order based algorithms fail.}
\end{itemize}

Both the multiple trajectory tracking and resistance to noise features have their limits. Dense random noise is easier to overcome as both the branch extension threshold and the minimum sequence length parameters can be increased to the point that it is extremely rare for unhelpful patterns to be learned. Structured noise on the other hand is more difficult to mitigate, as the interactions between multiple independent but repeating patterns can create exponentially many unreliable patterns that occur in high enough numbers to be stored within the EPST tree. Large numbers of stored patterns contribute to performance issues as described in Section \ref{EPSTTreeSizeProblem}.

Regarding the limitations of the experiments presented in this chapter, an obvious drawback lies in the simplicity of the synthetic datasets used. For this initial investigation, it was important to design simple scenarios to identify the broad features of the novel algorithm. An additional reason for the simplicity of the dataset is to compare the EPST to the sequential symbolic algorithms from which it was derived. The EPST can receive almost arbitrary event based data, including handling simultaneous events, which the sequential symbolic VMMs do not support. For these reasons, care was taken to provide a dataset that facilitated comparison.

It should be noted that concerning the implementation of the EPST, the counting method outlined in Section \ref{EPSTLearning} may lead to inconsistent counts in some situations. This may be remedied by methodical execution of the learning algorithm for each time step as opposed to the spike-triggered method. In practice, however, it is considered a testament to the robustness of the EPST that these situational counting errors are handled gracefully via the pruning extension, or by the reliability based selection method (Section \ref{EPSTRepresentativePattern}).

The EPST at its core produces an extremely complex hierarchical template over both time and event channels, and uses this complex template to decide whether to produce a prediction at a given time step. The templating system used in the EPST also contains a lot of redundancy, to which a lot of the robustness to noise may be attributed. The findings of this investigation illustrate that such a templating system can be extremely powerful in terms of coding capacity and robustness. Additional work is required to describe why the EPST is as powerful as it is in these respects, as well as to identify the limits of the computational features it exhibits. Biological neurons themselves may be found to use a high capacity spatiotemporal templating system similar to the EPST, in which case we could speculate this may be a reason for their robustness or computational power. Examples in the literature of individual biological neurons acting as complex spatiotemporal event filters were proposed in theory by Izhikevich \cite{polychronisation}. In biology, synaptic facilitation is explored in \cite{synapticmechanismsjackman2017}, and the extent to which dendritic morphology changes the responses of neurons to spikes is discussed in \cite{synapses_function} \cite{dendriteyang2014} \cite{dendritespruston2008}.

\subsection{Design Choices For Efficiency}
\label{EPSTEfficiency}

The EPST algorithm has shown promising results in event based prediction. However, the main drawback of the EPST algorithm is its high run time complexity, which can become problematic when dealing with datasets with a high density. In the worst case, if every $s$ length subsequence among the $N$ events in the window is already stored within the EPST tree, the algorithm tends to $O(\binom{N}{s})$. In practice, the true complexity is much lower than this upper bound, nevertheless the algorithm requires optimisation to reduce its runtime complexity.

\begin{figure}
	\centering
	\includegraphics[width=1.0\textwidth]{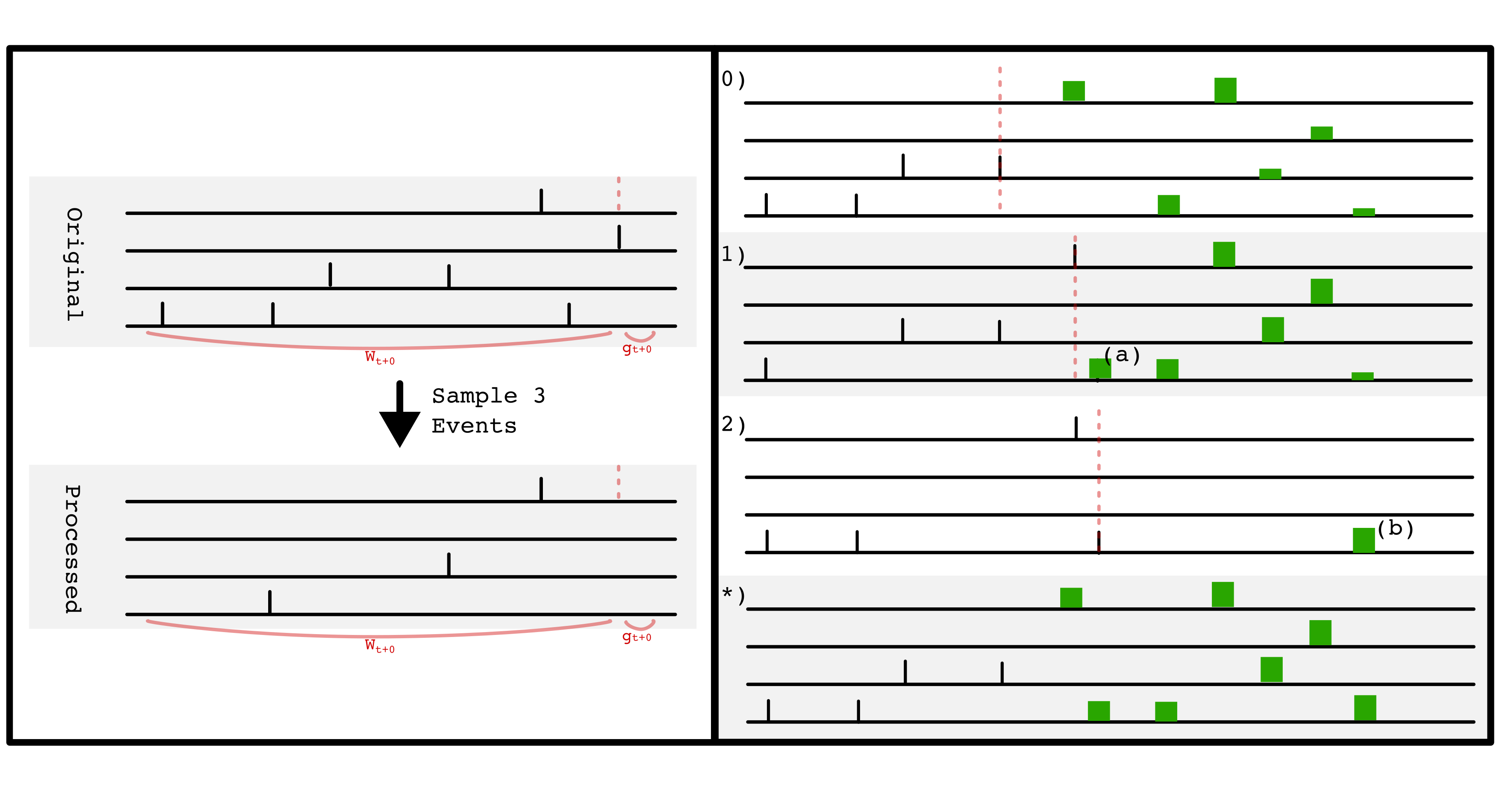}
	\caption{\label{CH5EPSTEfficiency} \textbf{A visual representation of the pre-prediction sampling optimisation.} The left panel, top, shows a collection of spikes in a context window $W_{t+0}$ to estimate the probability of a spike at time step $g_{t+0}$. The left panel, bottom, shows the same context after downsampling to 3 events. The right panel shows the effect of the sampling optimisation on the spike triggered prediction routine outlined in Figure \ref{EPSTDefinitionDiagram}. Instances 0, 1, and 2 show the result of 3 subsequent predictions, each with its own random sample from the context. The green bars visualise an example estimate spike probability for a time step, the larger the bar the higher the estimated probability. Instance (*) shows the aggregated estimate combining the three above estimates by calculating the max estimate for any given time step. Key points made by the illustration include a label (a) which shows an estimate that was missed in instance 0 due to the sampling optimisation removing the subsequence by chance, however, the following sample in instance 1 predicts the spike. Label (b) shows the only prediction from instance 2, which in this scenario increases an uncertain prediction made in prior steps. The fact that all other predictions are missed in the sample for instance 2 is not an issue, as prior instances have already committed values to these time steps.}
\end{figure}

To address this issue, before each prediction, a number of events can be uniformly randomly sampled from the context window without replacement. This can allow for faster runtime with minimal degradation in performance. By utilising this technique, the EPST algorithm can be optimised to provide predictions efficiently, even on datasets with high density. This method is illustrated in Figure \ref{CH5EPSTEfficiency}.

Another suggested optimisation method could be an early stopping threshold when matching stored patterns to the context window. In the non-optimised EPST, all matching subsequences are compared and collected, and the sequence with the highest reliability score is selected (High reliability score is low entropy, as explored in Section \ref{EPSTRepresentativePattern}). This can be made more efficient by stopping when a subsequence is found that has a sufficiently high reliability score. As the result of the full search is the selection of a single subsequence to represent the time window, this optimisation causes minimal loss in precision if the chosen threshold is sufficiently low.

\section{Conclusion}

From the computational neuroscience literature, many methods to represent the information held in a spike train have been investigated, all with associated advantages and disadvantages. Key issues exist with many modern spike based representations and the algorithms that use them. These issues include the requirement of a clearly defined global start or end to a spike based `code word', the necessity of sharing global information for routines such as normalisation, the requirement for offline learning, unrealistic assumptions about valid event patterns (such as disallowing simultaneous events), and a lack of resistance to certain types of noise. The goal of this study was to find an event based algorithm that addresses these issues.

The EPST presented in this study provides a flexible way to utilise the full representation space of event based data with minimal prior knowledge of the structure of the signal. This algorithm was shown to learn patterns quickly online under additive and subtractive noise conditions, outperforming established order based prediction algorithms (VMMs). The raw EPST has several drawbacks, such as a vulnerability to time jitter, as well as a large time and storage complexity. Hyperparameters, and extensions such as pruning and inhibition, were proposed to mitigate these drawbacks.

This study highlights the need to consider the unique strengths that event based data provides when designing algorithms. Matching sparse spatiotemporal patterns allows for a high capacity for representation and a robustness to error in recognition. Algorithms similar to the EPST fully utilise event based data sources, and should be developed to be more efficient while retaining flexibility and robustness. Finally, biological neurons should be investigated for computational properties similar to those of the EPST. Properties such as a high capacity for storing and matching spike patterns, the ability to distinguish multiple spatiotemporal patterns at once, or even the ability to resist random event noise to a greater extent than structured noise. If found, these computational properties may indicate that biological neurons utilise a complex spatiotemporal kernel to generate spikes.

\bibliographystyle{unsrtnat}

\end{document}